\documentclass[num-refs]{WileyMSP-template}
\usepackage{placeins}
\usepackage{amsmath,amssymb,amsfonts}
\usepackage{booktabs}
\usepackage{multirow}
\usepackage{nicematrix}
\usepackage{subcaption}
\usepackage{xcolor}
\usepackage{rotating}
\usepackage{svg}
\usepackage{url}
\usepackage{cite}
\usepackage{hyperref}
\usepackage{orcidlink}
\usepackage{textcomp}
\usepackage{etoolbox}

\newcommand\modification[1]{\textcolor{black}{#1}}

\begin{document}

\pagestyle{fancy}
\rhead{\includegraphics[width=2.5cm]{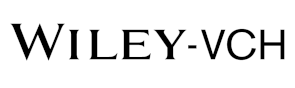}}

\title{Performance evaluation of super‑resolution techniques for characterization and phenotyping of commodity crops using multispectral imagery}

\maketitle


\author{Abhiram Karukayil$^{1}$\orcidlink{0009-0000-4208-6701}, Jo\~ao F.\ C.\ Mota$^{1}$\orcidlink{0000-0001-7263-8255}, Mateo Vásquez-Guevara$^{2}$\orcidlink{0009-0001-3333-8142}, Tito Arevalo-Ramirez$^{2}$\orcidlink{0000-0003-2542-6545} and Fernando Auat Cheein$^{3,4}$\orcidlink{0000-0002-6347-7696}}

\begin{affiliations}
$^{1}$ \quad School of Engineering and Physical Sciences, Heriot-Watt University, Edinburgh, UK, ak5001@hw.ac.uk, j.mota@hw.ac.uk\\
$^{2}$ \quad Department of Electrical Engineering, Department of Mechanical Engineering and Metallurgy, School of Engineering, Faculty of Engineering, Pontificia Universidad Católica de Chile, Santiago, Chile; mvasquezg9@estudiante.uc.cl, tito.arevalo@uc.cl\\
$^{3}$ \quad School of Mathematical and Computer Sciences, Heriot-Watt University, Edinburgh, UK\\
$^{4}$ \quad Harper Adams University, England, UK, fauat@harper-adams.ac.uk\\
\medskip
\textbf{Correspondence:} Jo\~ao F.\ C.\ Mota (j.mota@hw.ac.uk)

\end{affiliations}










\begin{abstract}
\modification{Multispectral imagery provides spectral information beyond the visible range for vegetation monitoring, but it typically suffers from limited spatial resolution, which can obscure fine crop structures and affect vegetation-index (VI) estimates.} This study evaluates whether single-image super-resolution (SISR) methods developed for red, green, and blue (RGB) imagery can enhance the spatial resolution of red, green, and near-infrared (RGN) crop images while preserving spectral fidelity. Eight pretrained methods are applied directly to RGN images without fine-tuning: bicubic and nearest-neighbor interpolation, Real-ESRGAN, HAT, HiT-SR, SRFormer, ResShift, and StableSR. Reconstruction at $4\times$ and $16\times$ is evaluated for whole images and manually segmented leaf regions. Reconstruction quality is assessed using mean absolute error (MAE), peak signal-to-noise ratio (PSNR), structural similarity index (SSIM), spectral angle mapper (SAM), and erreur relative globale adimensionnelle de synthèse (ERGAS). Eight VIs, computed from reference and super-resolved images, are evaluated using MAE, PSNR, and SSIM. Transformer-based methods \modification{significantly} outperform GAN- and diffusion-based approaches in both evaluation settings \modification{($p < 0.001$)}, with \modification{HAT, HiT-SR, and SRFormer forming the highest-performing statistical tier}. These findings indicate that RGB-trained SISR can enhance spatial detail in RGN crop imagery while maintaining spectral consistency and VI reliability for high-resolution crop phenotyping.
\end{abstract}


\section{Introduction}
The increasing global demand for food, coupled with the overuse of arable land and climate instability, has accelerated the adoption of precision agriculture. A key technology facilitating this transition is multispectral remote sensing, which allows non-destructive monitoring of crop canopies over large areas, either from unmanned aerial vehicles (UAVs) or from satellite platforms~\cite{jiang2023combining}. Vegetation indices (VIs) such as the normalized difference vegetation index (NDVI) and the optimized soil-adjusted vegetation index (OSAVI) are commonly derived from multispectral bands to quantify canopy vigor, biomass, and structural characteristics~\cite{mwinuka2022uav}. However, the relatively low spatial resolution of multispectral sensors, which prioritize spectral fidelity over spatial detail, limits their ability to resolve fine-scale structures such as individual leaves. As a result, pixels capturing both healthy and stressed tissues can distort spectral responses and reduce the reliability of VIs for diagnostic applications~\cite{de2024assessing}.

Image super-resolution (SR) has emerged as a promising approach to address these spatial limitations by reconstructing high-resolution images from low-resolution observations. Super-resolution methods are broadly categorized into single-image super-resolution (SISR) and multi-frame super-resolution (MFSR)~\cite{liu2021research}. SISR methods reconstruct high-resolution images from a single input by exploiting internal image statistics, whereas MFSR methods combine information from multiple observations of the same scene. Traditional SISR approaches include classical techniques such as nearest-neighbor upsampling and bicubic upsampling, while recent advances in deep learning have led to substantial improvements using convolutional neural networks (CNNs)~\cite{tian2022heterogeneous}, generative adversarial networks (GANs)~\cite{wang2021real}, transformers~\cite{zhou2023srformer}, and diffusion models~\cite{yue2023resshift}. These models learn from large datasets strong image priors and long-range dependencies, achieving state-of-the-art performance on red, green, and blue (RGB) image reconstruction tasks. Previous studies  demonstrated the potential of SR in agricultural contexts, such as disease monitoring and UAV-based crop analysis~\cite{palan2025leveraging,yamamoto2017super,wu2022wheat}. However, most existing SR architectures are designed for RGB images and optimized for perceptual quality rather than pixel fidelity~\cite{Vella21-RobustSingleImageSuperResolutionViaCNNsAndTVTVMinimization}. This raises concerns about their suitability for multispectral data, where spectral accuracy is a key factor for reliable vegetation analysis.

\modification{Multispectral and hyperspectral imaging captures spatial and spectral information simultaneously, providing detailed spectral measurements for remote-sensing applications~\cite{hong2026hyperspectral}. Building on these capabilities, super-resolution methods have been developed to enhance the spatial resolution of both types of imagery. Recent unsupervised methods, including M2U-Net~\cite{li2024model}, EDIP-Net~\cite{li2025enhanced}, and XINet~\cite{li2023x}, investigate hyperspectral reconstruction using image observations, spectral information, and image-specific priors. Related work has also explored spectral-band modelling for SR and large-scale representation learning for remote sensing, as illustrated by MambaX~\cite{li2025mambax} and SpectralGPT~\cite{hong2024spectralgpt}. These studies are designed for hyperspectral or multispectral data, whereas the present work examines the direct transfer of RGB-trained SISR methods to RGN crop imagery.} In agricultural applications, UAV-based multispectral imaging can acquire data from much shorter distances than satellite platforms, but payload capacity, swath width, sensor optics, and operational constraints can still limit the spatial resolution obtainable for fine-scale crop analysis. High-resolution multispectral sensors may also impose additional cost, weight, and power requirements~\cite{phang2023satellite}. Consequently, SR may offer a practical alternative for enhancing spatial detail in multispectral imagery, provided that spectral integrity and vegetation-index reliability are preserved. Since this study focuses exclusively on single-image super-resolution, the terms SR and SISR are used interchangeably here.

Motivated by these challenges, we investigate the applicability of RGB-trained SISR models for multispectral crop images, with a focus on improving spatial resolution. To assess their ability to transfer beyond the RGB domain, we apply the methods directly to multispectral RGN images. \modification{All super-resolution methods are evaluated using their publicly available pretrained implementations and default inference configurations, without adapting the models to the acquired RGN imagery, allowing us to compare their performance when directly applied to RGN multispectral imagery.} In the absence of UAV deployment, and to emulate low-altitude observations, we acquired RGN imagery at ground level using a MAPIR Survey3W multispectral camera, \modification{resulting in a horizontal or near-horizontal viewing perspective.} We consider eight SISR methods, spanning four methodological categories: classical filtering approaches (bicubic and nearest neighbor~\cite{fattal2007image}), GAN-based (Real-ESRGAN~\cite{wang2021real}), transformer-based (hybrid attention transformer (HAT)~\cite{chen2025hat}, hierarchical transformer for efficient image super-resolution (HiT-SR)~\cite{zhang2024hit}, SRFormer~\cite{zhou2023srformer}), and diffusion-based (ResShift~\cite{yue2023resshift}, StableSR~\cite{wang2024exploiting}). To provide a comprehensive benchmark, we also consider two scale factors, $4\times$ and $16\times$, and analyze two different scenarios: whole-image reconstruction, where metrics are computed across the entire image, and leaf-level reconstruction, where the same images are evaluated only within manually segmented leaf regions using masked analysis. \modification{For the 4$\times$ experiments, each image was processed once using the corresponding 4$\times$ model. For 16$\times$ reconstruction, the same 4$\times$ inference procedure was applied sequentially twice. This protocol was consistently applied across all methods and acquired images.} Reconstruction performance on RGN images is assessed using mean absolute error (MAE)~\cite{hodson2022root}, peak signal-to-noise ratio (PSNR)~\cite{hore2010image}, structural similarity index (SSIM)~\cite{hore2010image}, spectral angle mapper (SAM), and erreur relative globale adimensionnelle de synthèse (ERGAS). And VI evaluation is performed using eight indices computed from both reference and super-resolved images, where performance is measured on the resulting VI maps using MAE, PSNR, and SSIM. Our main contribution is a systematic benchmark of RGB-trained SR methods for multispectral crop imagery, examining how different SR architectures influence spatial reconstruction quality, spectral consistency, and VI behavior for precision crop phenotyping.

\section{Materials and Methods}
This section describes how we prepared the dataset, implemented the super-resolution (SR) methods, and used vegetation indices (VI) in the experiments.
\subsection{Data Acquisition}
\begin{figure}[t]
\centering
\includegraphics[width=.87\linewidth]{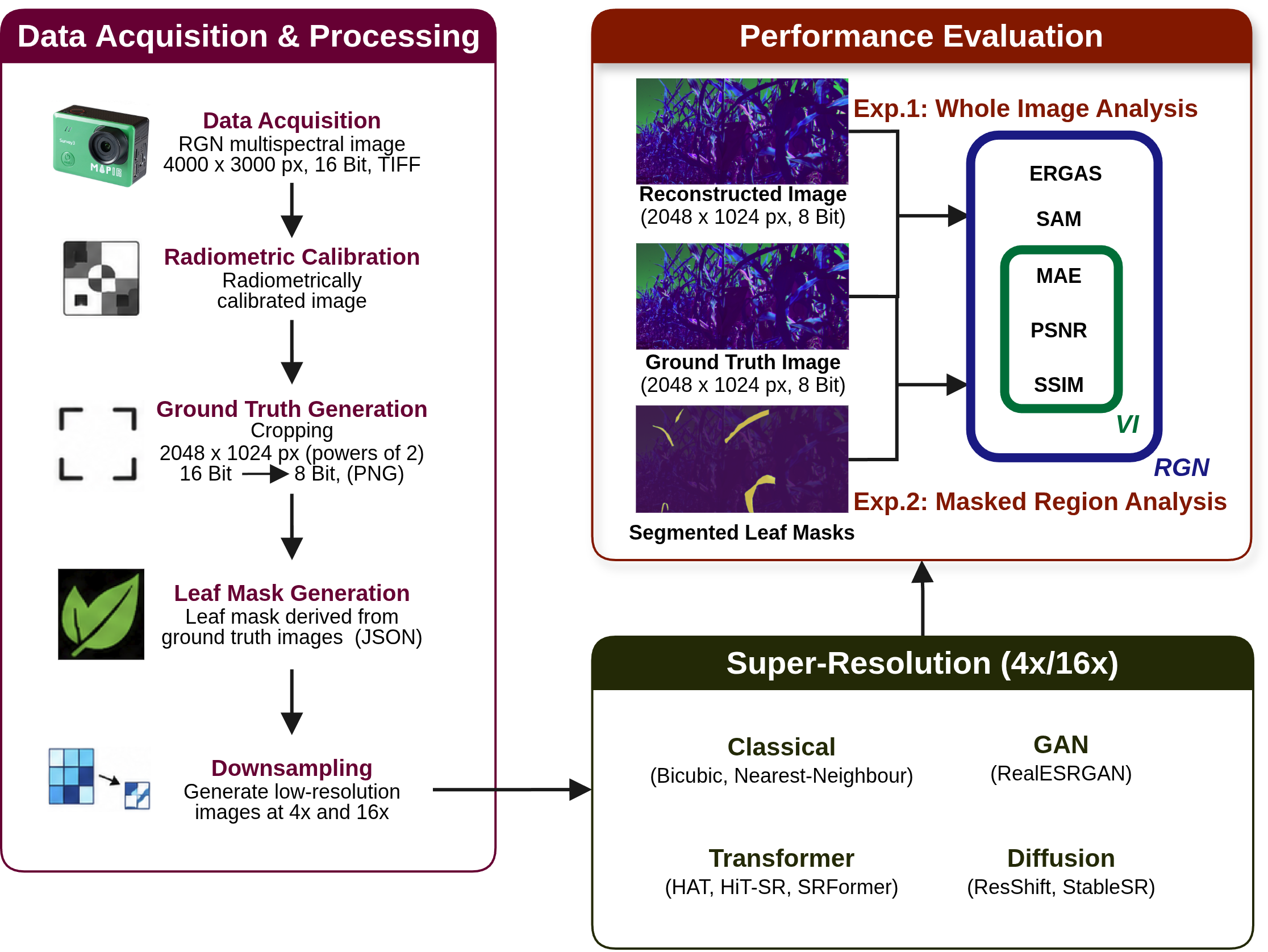}
\caption{Our experimental workflow for multispectral image acquisition, super-resolution processing, and performance assessment through whole-image (Experiment 1, Sec.~\ref{sec:ex1}) and leaf-level (Experiment 2, Sec.~\ref{sec:ex2}) evaluation. Groundtruth images are generated during data acquisition \& processing by extracting a centered $2048 \times 1024$ image patch and converting it from 16-bit to an 8-bit portable network graphics (PNG) format. The images are subsequently downsampled by factors of $\times4$ and $\times16$, reconstructed using different SR methods, and evaluated at the whole-image and masked-region levels. The RGN reconstruction metrics are indicated in the blue box, while VI reconstruction metrics are indicated in the green box.}

\label{fig:work_flow}
\end{figure}
\begin{figure}[t]
    \centering
    \begin{subfigure}[t]{0.32\textwidth}\centering\includegraphics[width=\linewidth]{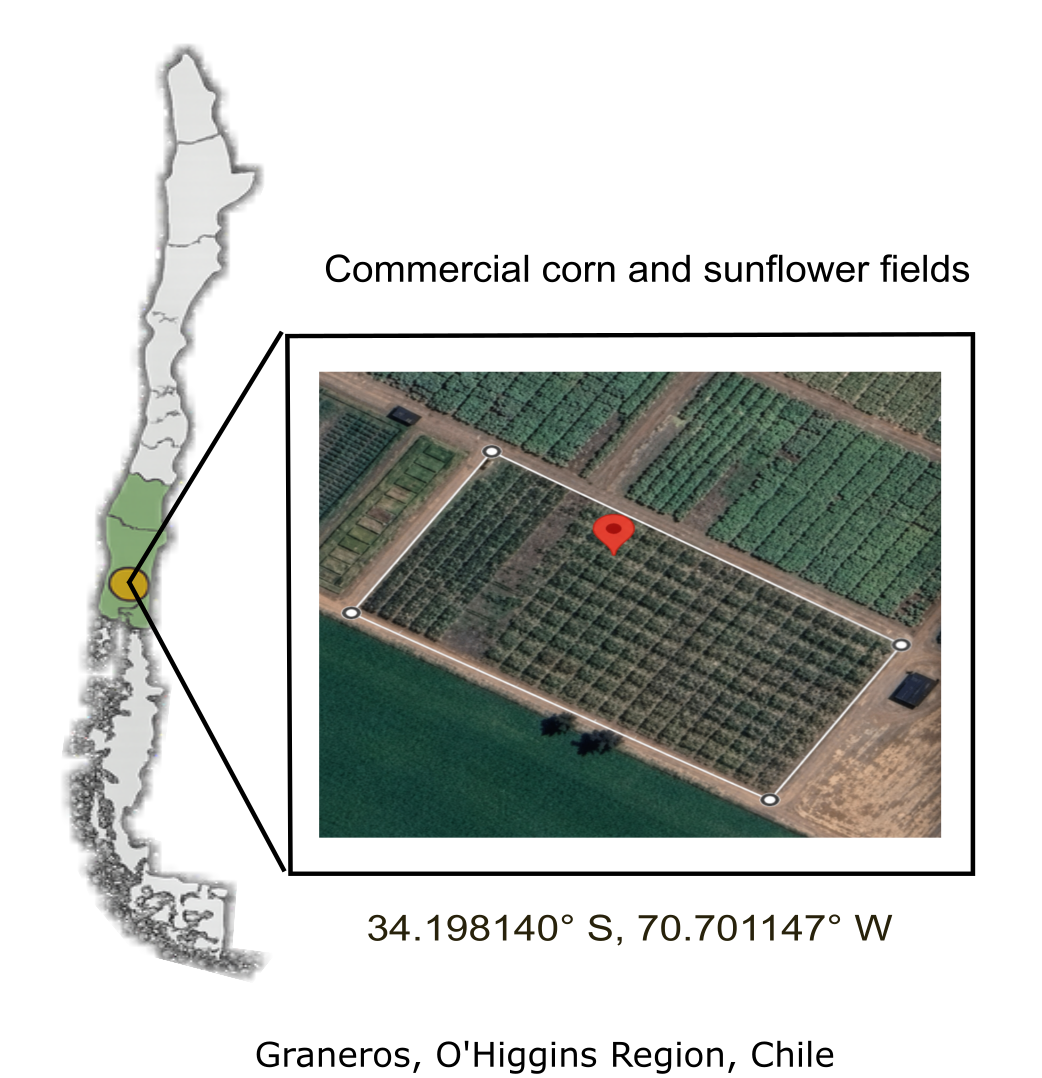}
    \caption{\label{subfig:setupmap}Study area and field location}
    \end{subfigure}
    \hfill
    \begin{subfigure}[t]{0.32\textwidth}\centering\includegraphics[width=\linewidth]{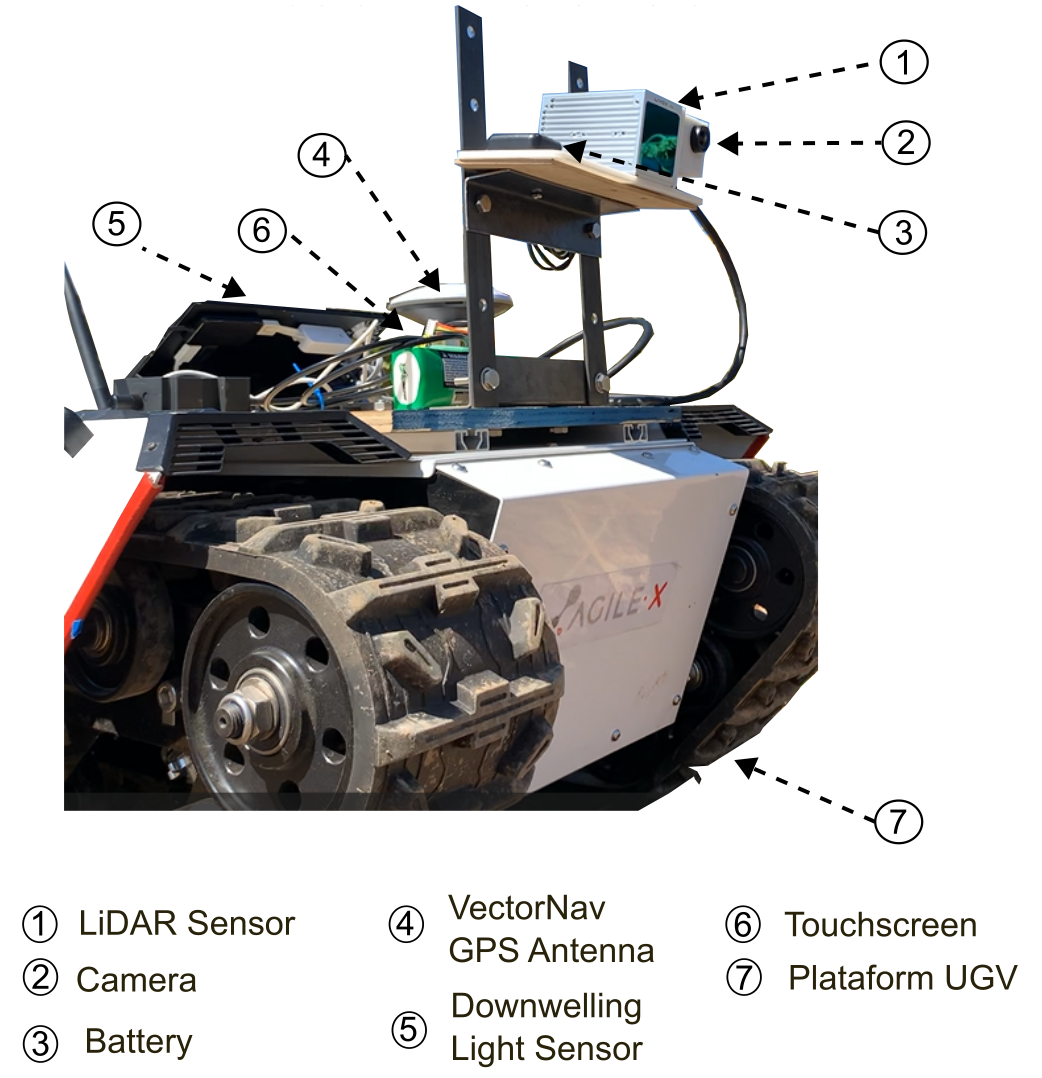}\caption{\label{subfig:setupugv}UGV platform}
    \end{subfigure}
    \hfill
    \begin{subfigure}[t]{0.32\textwidth}\centering\includegraphics[width=\linewidth]{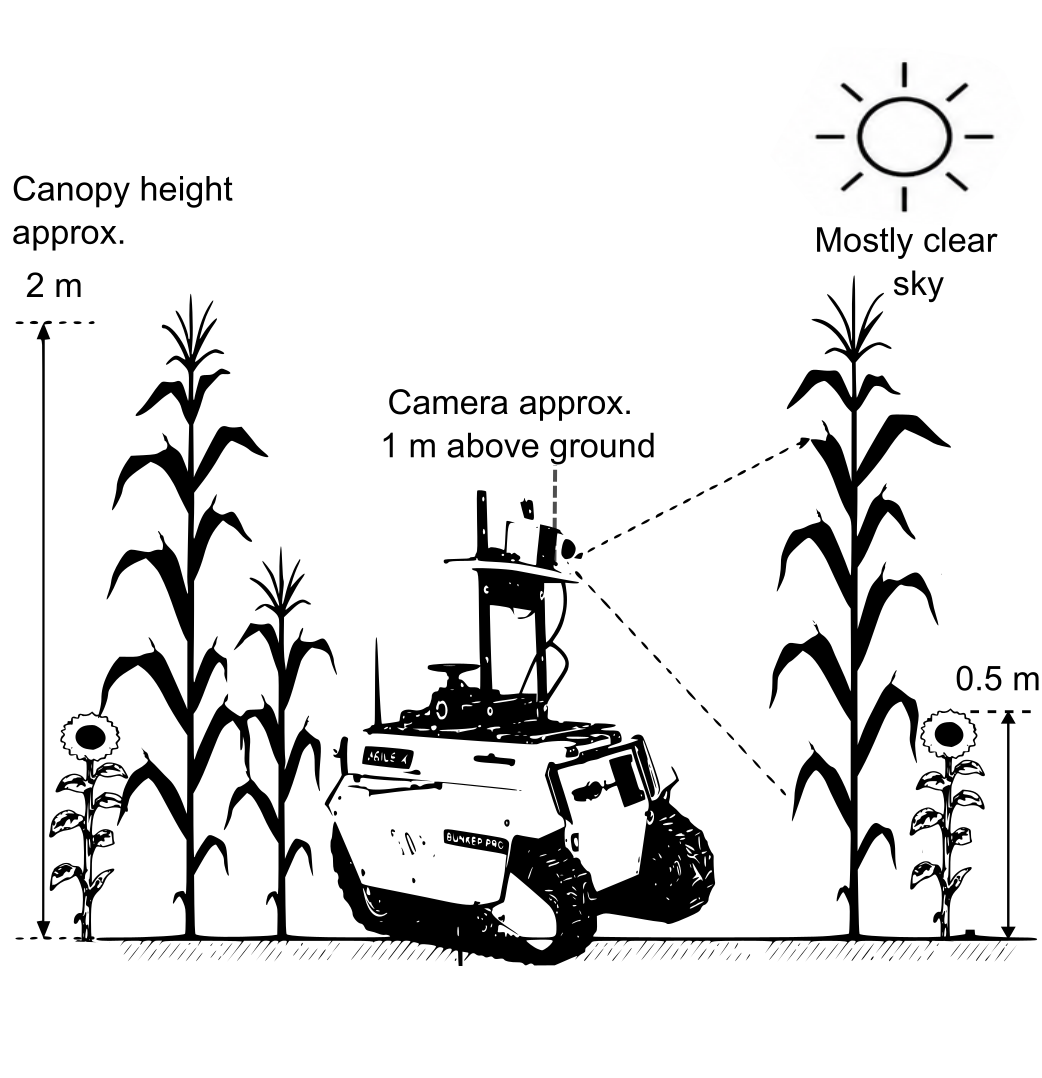}\caption{\label{subfig:setupplants}Multispectral acquisition setup}
    \end{subfigure}
    \caption{
    Ground-based multispectral data acquisition framework. 
    (a) Study area and field location. 
    (b) Bunker Agilex UGV platform configuration. 
    (c) Ground-based multispectral acquisition setup.
    }
    \label{fig:acquisition}
\end{figure}

We collected the data between 11:00 and 13:00 UTC$-4$ on 7 March 2025 in a commercial corn field located at 34.198140$^\circ$ S, 70.701147$^\circ$ W, in Graneros commune, Cachapoal Province, Libertador General Bernardo O'Higgins Region, Chile; see Fig.~\ref{subfig:setupmap}. The study area is characterized by a Mediterranean climate, with cold, partly cloudy winters and warm, dry summers, and reported annual temperatures ranging approximately from \(4^\circ\)C to \(29^\circ\)C \cite{weather1}. During the acquisition period, weather conditions were mostly clear, with air temperatures ranging from 22.2 to 24.5~$(^\circ)$C, relative humidity between 37\% and 46\%, wind speeds of 5.0--11.4~km~h\(^{-1}\), cloud cover of 1--7\%, and shortwave solar radiation between 731 and 866~W~m\(^{-2}\) \cite{openmeteo2}. 

\begin{figure}[t]
    \centering
    \begin{subfigure}{0.45\linewidth}
\centering\includegraphics[width=\linewidth]{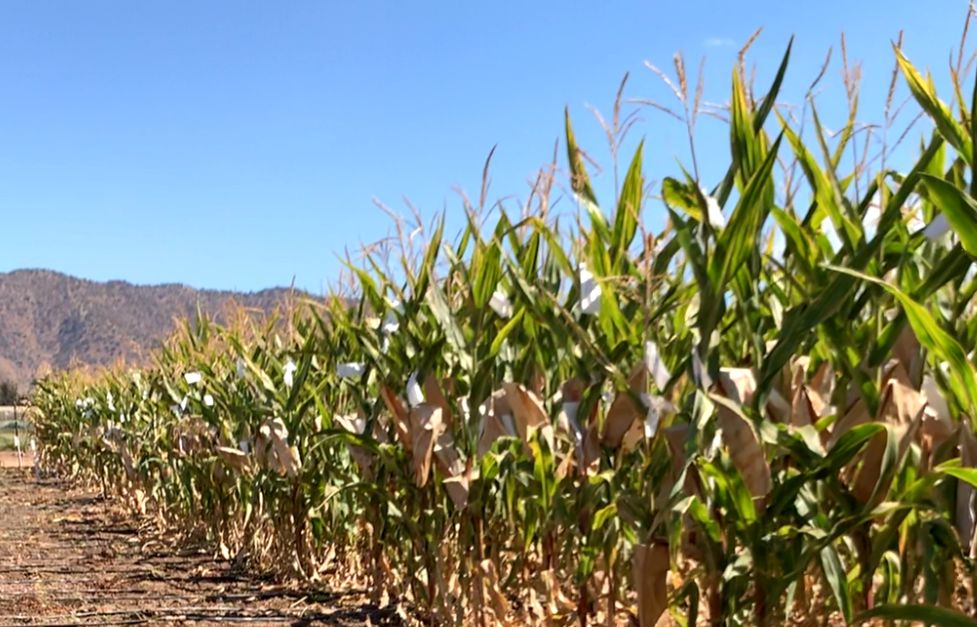}
        \caption{Maize}
    \end{subfigure}
    \hspace{0.4cm}
    \begin{subfigure}{0.45\linewidth}
        \centering        \includegraphics[width=\linewidth]{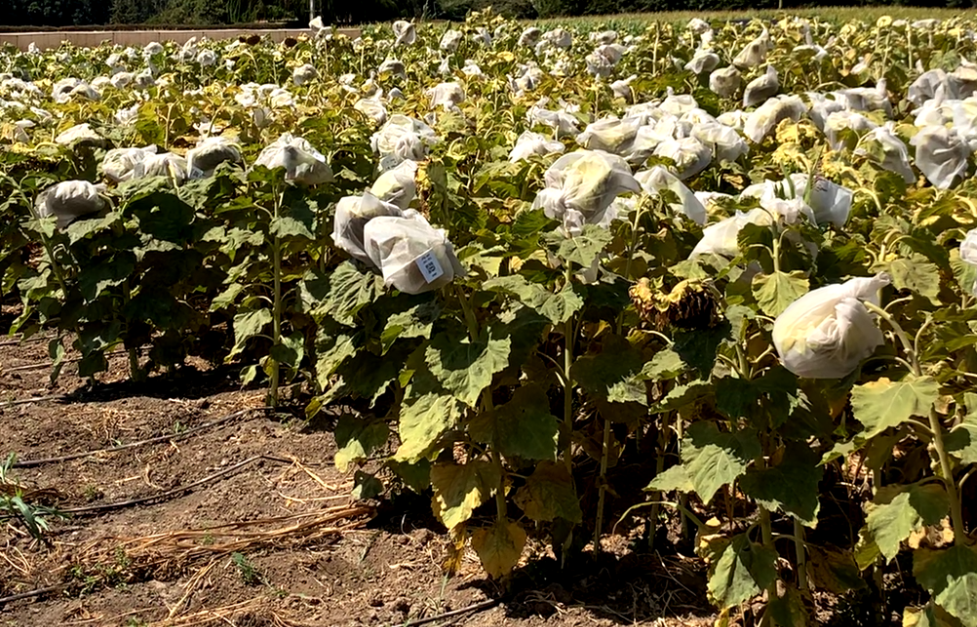}
        \caption{Sunflower}
    \end{subfigure}
    \caption{RGB images of (a) maize and (b) sunflowers during data collection under identical imaging conditions.}
    \label{fig:crop_rgb}
\end{figure}

At the time of acquisition, the maize crop was at the V14 phenological stage, approximately ten days before flowering, with an average canopy height of 2~m. In contrast, the sunflower crop was at approximately 10\% flowering, with an average canopy height of 0.5~m. The field corresponded to experimental maize and sunflower material sown on 30 October 2024, with a row spacing of 0.70~m and an approximate plant density of 70{,}000 plants~ha\(^{-1}\). See Fig.~\ref{fig:crop_rgb} for illustrative RGB images. We acquired multispectral data using a MAPIR Survey3W multispectral camera mounted on a Bunker Agilex unmanned ground vehicle (UGV), positioned approximately 1~m above the ground, as illustrated in Figs~\ref{subfig:setupugv}-\ref{subfig:setupplants}. The camera captures red, green, and near-infrared bands centered at 660, 550, and 850~nm, respectively, enabling the computation of vegetation indices such as the normalized difference vegetation index (NDVI). The system is equipped with a Sony Exmor R IMX117 12~MP CMOS sensor and produces images with a spatial resolution of \(4000 \times 3000\) pixels. In addition, the camera incorporates a non-fisheye wide-angle lens, which reduces geometric distortion and preserves spatial image quality for subsequent analysis. Images were acquired with ISO fixed at 50, aperture set to \(f/2.8\), and a focal length of 3.37~mm, while exposure time was automatically adjusted according to illumination conditions, ranging approximately from \(1/447\) to \(1/133\)~s during field acquisition. RAW and JPEG image pairs were recorded simultaneously at intervals of approximately 2--3~s. 
While RAW files preserve 12-bit per-channel radiometric information required for quantitative analysis, JPEG files provide 8-bit visual references for rapid inspection.


\subsection{Data Preparation}

Figure~\ref{fig:work_flow} shows our experimental workflow, from data acquisition to evaluation. Following the manufacturer's recommendation, we performed radiometric calibration using MAPIR camera control software \cite{MAPIR} and used reflectance calibration targets acquired during the same experimental session. Specifically, we used the MAPIR diffuse reflectance standard calibration target package T4-R125, which contains four diffuse reflectance standards of known reflectance, each measuring \(125 \times 125\)~mm, together with a QR/fiducial marker for automatic target detection and panel sampling. According to the manufacturer, the reflectance standards are characterized over the 250--2500~nm spectral range using spectrophotometers equipped with an integrating sphere. We used 17 calibration-target images within the MAPIR Camera Control workflow: 13 (resp.\ 4) images acquired before (resp.\ after) the field sequence. The RAW images were imported into the software, the T4-R125 calibration target was automatically identified through the QR/fiducial marker, and the four reflectance panels were sampled to estimate the relationship between pixel intensity and surface reflectance. MAPIR Camera Control then applied the radiometric calibration to the RAW image sequence and exported the results as reflectance-calibrated, 16-bit, three-band tagged image file format (TIFF) images. This workflow reduces variability associated with illumination changes, sensor spectral response, and optical effects such as vignetting, thereby improving the consistency of reflectance measurements across the dataset.

\begin{figure}[h]
\centering
\includegraphics[width=.55\linewidth]{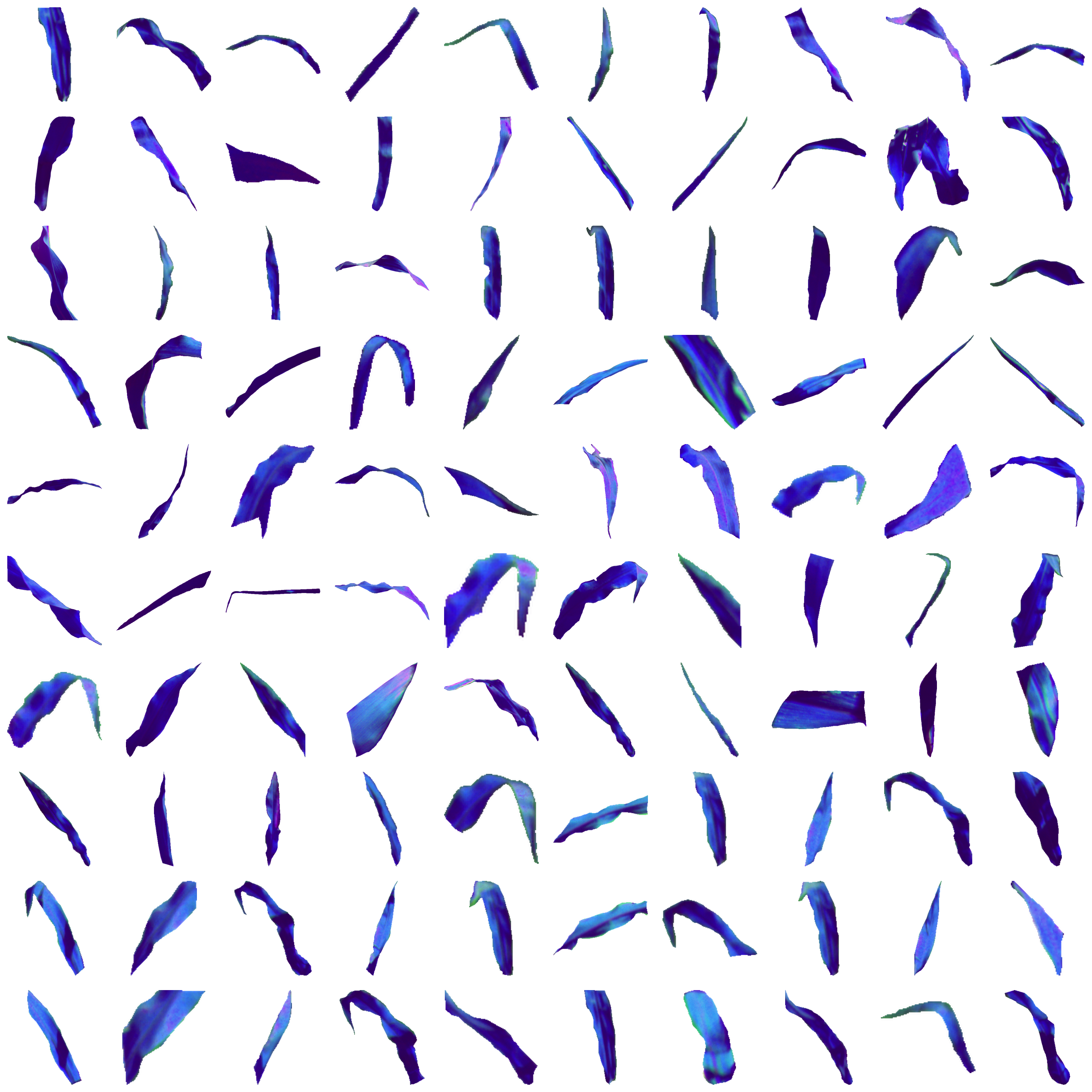}
\caption{Visualization of the 100 manually segmented maize leaves  used for masked-region analysis (Sec.~\ref{sec:ex2}). Note the  morphological variability within the dataset.}

\label{fig:leaf_collage}
\end{figure}

To evaluate the performance of SISR techniques, we acquired a dataset of 75 radiometrically calibrated multispectral images comprising red, green, and near-infrared (RGN) bands. The dataset included 35 maize field images and 40 sunflower field images, which served as the high-resolution reference data for all experiments. Groundtruth images were generated by cropping the original 4000$\times$3000 pixel, 16-bit TIFF images to 2048×1024 pixels, ensuring compatibility with the $4\times$ and $16\times$ downsampling and reconstruction procedures. Low-resolution inputs were generated by downsampling the reference images using the INTER\_AREA method implemented in OpenCV~\cite{bradski2000opencv} in Python. The cropped images were subsequently converted to 8-bit PNG format for compatibility with all the evaluated SR methods. We used the full dataset to analyze whole-image reconstruction (Sec.~\ref{sec:ex1}), and 100 maize leaves for leaf-level reconstruction (Sec.~\ref{sec:ex2}). For the latter, we selected 20 non-adjacent images, from which we manually segmented 5 leaves using labelme-ytu~\cite{labelmeytu}. The resulting annotations were stored in javascript object notation (JSON) format and used for the masked-region evaluation, as shown in Figure~\ref{fig:leaf_collage}.


\begin{table*}[th]
    \centering
    \renewcommand{\arraystretch}{1.2}
    \caption{Super-resolution algorithms evaluated in this study. They are classified by category and, for the learning-based methods, the training datasets are also indicated.}
    \vspace{-0.2cm}
    \begin{tabular}{l l l} 
        \toprule
         \textbf{SR Technique}& \textbf{Category} & \textbf{Training Datasets}\\ 
         \midrule
          Bicubic \cite{khaledyan2020low} & Classical& -\\
           NN \cite{fattal2007image} & Classical& -\\
          HAT \cite{chen2025hat} & Transformer& ImageNet~\cite{deng2009imagenet} and DF2K( DIV2K $+$ Flickr2K~\cite{timofte2017ntire}) \\
          HiT-SR~\cite{zhang2024hit}& Transformer& DIV2K~\cite{agustsson2017ntire}\\
            SR Former~\cite{zhou2023srformer} &Transformer&DIV2K~\cite{agustsson2017ntire} and DF2K( DIV2K $+$ Flickr2K~\cite{timofte2017ntire})\\
          Real ESRGAN~\cite{wang2021real}& GAN & DIV2K~\cite{agustsson2017ntire}, Flickr2K~\cite{timofte2017ntire} and OutdoorSceneTraining (OST)~\cite{wang2018recovering}\\
          ResShift~\cite{yue2023resshift, yue2024efficient}& Diffusion Model& ImageNet~\cite{deng2009imagenet}\\

          Stable SR~\cite{wang2024exploiting} &Diffusion Model& DIV2K~\cite{agustsson2017ntire}, DIV8K~\cite{gu2019div8k}, Flickr2K~\cite{timofte2017ntire}, OST~\cite{wang2018recovering} and FFHQ~\cite{karras2019style} \\
         \bottomrule
    \end{tabular}
    \label{tab:srt}
    \end{table*}

\subsection{Super-Resolution Methods} \label{sec:SR}

Table~\ref{tab:srt} summarizes the SR methods considered in this study. They include classical interpolation, GAN-based, transformer-based, and diffusion-based, providing a representative spectrum of modern SR approaches. Bicubic interpolation~\cite{khaledyan2020low} estimates new pixel values using cubic interpolation in a local neighborhood, based on a $4\times4$ grid of neighboring pixels. Nearest-neighbor (NN) interpolation~\cite{wang2020deep} assigns to each output pixel the value of the closest input pixel without introducing additional spatial information. Although NN interpolation does not improve image detail, it preserves the original pixel arrangement, although this often results in blocky artifacts and reduced visual smoothness. Transformer-based methods, such as hybrid attention transformer (HAT)~\cite{chen2025hat}, hierarchical transformer for efficient image super-resolution (HiT-SR)~\cite{zhang2024hit}, and SRFormer~\cite{zhou2023srformer}, use self-attention mechanisms to capture long-range spatial dependencies with the aim of recovering fine structural and textural details. They overcome limitations of early vision transformers by improving cross-window and multi-scale feature interactions. For example, HAT~\cite{chen2025hat} enhances information exchange via hybrid attention, which combines channel attention with overlapping spatial attention. HiT-SR~\cite{zhang2024hit}, on the other hand, employs a hierarchical multi-scale windowing strategy with efficient spatial–channel interaction modules to capture broader contextual dependencies with reduced computational complexity. SRFormer implicitly expands the receptive field by using a permuted self-attention mechanism. Real-ESRGAN~\cite{wang2021real} builds on the generative adversarial network (GAN) model ESRGAN~\cite{wang2018esrgan} by incorporating a high-order degradation model to simulate realistic complex distortions. These methods are trained using adversarial and perceptual losses, which encourage visually sharp and perceptually realistic textures, albeit at the cost of reduced pixel-level accuracy. Training stability is improved by using a U-Net-based~\cite{lu2022single} discriminator with spectral normalization, which enhances the discrimination of structural artifacts and stabilizes adversarial learning. Diffusion-based methods, such as ResShift~\cite{yue2023resshift, yue2024efficient} and StableSR~\cite{wang2024exploiting}, formulate SR as an iterative denoising process. These models progressively refine low-resolution inputs through stochastic sampling steps. However, reliance on iterative refinement may introduce deviations in spectral accuracy in remote sensing applications where radiometric consistency must be maintained.

\modification{Overall, the methods assessed here mainly differ in how they reconstruct high-resolution information from low-resolution inputs. Traditional interpolation techniques estimate pixel values using information from neighboring pixels without learning from data, making them straightforward and computationally efficient. Transformer-based approaches learn spatial features and use attention mechanisms to model dependencies across different regions of an image, supporting the reconstruction of complex structures. GAN-based methods use adversarial training and perceptual losses to produce more visually realistic outputs, although the synthesized details may not closely correspond to the ground-truth pixel values. Diffusion-based techniques formulate reconstruction as an iterative denoising process, enabling the generation of fine details but with higher computational cost and the potential to introduce artifacts.}

\subsection{Vegetation Indices} \label{sec:Vegetation_indices}

Vegetation indices (VIs), which enable assessing crop condition, biomass, canopy vigor, and physiological stress, can be computed from  reflectance values obtained from multispectral images. Table~\ref{tab:paperVI} summarizes the VIs we used, all of which computed from the red, green, and near-infrared (NIR) bands. The normalized difference vegetation index (NDVI) and difference vegetation index (DVI) were used to characterize vegetation presence and canopy condition. Biomass-related characteristics were assessed using the ratio vegetation index (RVI) and modified simple ratio (MSR), both based on the relationship between NIR and red reflectance. The MSR can reduce saturation effects in dense vegetation. To account for soil background effects, we also evaluated the soil adjusted vegetation index (SAVI) and optimized soil adjusted vegetation index (OSAVI). Physiological stress and chlorophyll-related responses were assessed using the green normalized difference vegetation index (GNDVI), which uses green reflectance instead of red reflectance and is more sensitive to variations in chlorophyll content.

\renewcommand{\arraystretch}{2.2}
\setlength{\extrarowheight}{2pt}

\begin{table*}[th]
\centering
\caption{Vegetation indices used in this study to assess super-resolution reconstruction. 
$R_{\text{NIR}}$, $R_{\text{Red}}$, and $R_{\text{Green}}$ denote reflectance
values of the near-infrared (850~nm), red (550~nm), and green (660~nm) bands.}
\vspace{-0.1cm}
\renewcommand{\arraystretch}{1.1}
\begin{tabular}{llp{7.7cm}c}
\toprule
\textbf{Index} & \textbf{Description} & \textbf{Formula} & \textbf{Ref.} \\
\midrule
DVI & Difference Vegetation Index 
& $\displaystyle R_{\text{NIR}} - R_{\text{Red}}$
& \cite{tucker1979red} \\
MSAVI & Modified Soil Adjusted Vegetation Index 
& $\displaystyle 2R_{\text{NIR}} + \frac{1}{2} - \frac{1}{2}\sqrt{(2R_{\text{NIR}} + 1)^2 - 8(R_{\text{NIR}} - R_{\text{Red}})}$
& \cite{he2015critical} \\
NDVI & Normalized Difference Vegetation Index 
& $\displaystyle (R_{\text{NIR}} - R_{\text{Red}})/(R_{\text{NIR}} + R_{\text{Red}})$
& \cite{gitelson2002novel} \\
OSAVI & Optimized Soil Adjusted Vegetation Index 
& $\displaystyle 1.16(R_{\text{NIR}} - R_{\text{Red}})/(R_{\text{NIR}} + R_{\text{Red}} + 0.16)$
& \cite{rondeaux1996optimization} \\
RVI & Ratio Vegetation Index 
& $\displaystyle R_{\text{NIR}}/R_{\text{Red}}$
& \cite{bannari1995review} \\
SAVI & Soil Adjusted Vegetation Index 
& $\displaystyle 1.5(R_{\text{NIR}} - R_{\text{Red}})/(R_{\text{NIR}} + R_{\text{Red}} + 0.5)$
& \cite{huete1988soil} \\
GNDVI & Green Normalized Difference Vegetation Index 
& $\displaystyle (R_{\text{NIR}} - R_{\text{Green}})/(R_{\text{NIR}} + R_{\text{Green}})$
& \cite{gitelson1996use} \\
MSR & Modified Simple Ratio 
& $\displaystyle (R_{\text{NIR}}/R_{\text{Red}} - 1)/\sqrt{R_{\text{NIR}}/R_{\text{Red}} + 1}$
& \cite{wu2008estimating} \\
\bottomrule
\end{tabular}
\label{tab:paperVI}
\end{table*}


\section{Experimental Results}

\modification{This section presents the evaluation metrics and assesses the performance of the super-resolution methods described in Section~\ref{sec:SR}. Their ability to preserve spatial detail and spectral consistency is evaluated using the complementary metrics introduced in Section~\ref{sec:metrics}. Since multispectral super-resolution (SR) requires the preservation of both spatial detail and spectral information, reconstruction quality is evaluated from these two perspectives. For readability, all MAE values are scaled by 100, and the input RGB images are normalized to the range $[0,1]$.}

We conducted two complementary experiments to evaluate reconstruction performance at different spatial scales. Experiment~1, reported in Section~\ref{sec:ex1}, evaluates whole-image reconstruction using the complete dataset of 35 maize-field images and 40 sunflower-field images. Experiment~2, reported in Section~\ref{sec:ex2}, evaluates leaf-level reconstruction within manually segmented mask regions using 100 maize leaves extracted from the same dataset. Together, these experiments allow us to assess the performance of the competing approaches both at the whole-image level and within individual leaf regions. In both experiments, reconstruction quality for RGN images is measured using MAE, PSNR, SSIM, SAM, and ERGAS, while VI reconstruction quality is evaluated using MAE, PSNR, and SSIM. Numerical values reported below in the text or in tables are average values computed across all the data samples. In contrast, boxplots visualize the distribution of errors and display the median as their central line. The evaluation metrics used in these experiments are described below.

\subsection{\modification{Evaluation Metrics}} \label{sec:metrics}

Unlike RGB or grayscale images, multispectral SR requires preserving both spatial detail and spectral information. To evaluate reconstruction performance, we thus selected a range of metrics, each of which captures a different aspect of image quality. Given an image of size $H \times W \times C$, where $H$ and $W$ denote spatial dimensions and $C$ the number of channels/bands, we represent the value of each pixel of the reconstructed (resp.\ groundtruth) image by $x_i \in \mathbb{R}$ (resp.\ $y_i \in \mathbb{R}$), for $i = 1, \ldots, N := H\cdot W\cdot C$. We consider $C=1$ for single-channel VI analysis and $C=3$ for RGN images.

To assess reconstruction quality, we used the mean absolute error (MAE)~\cite{hodson2022root}, 
$(1/N)\sum_{i=1}^{N}\left|x_i-y_i\right|$,
with lower values indicating smaller reconstruction errors and hence better performance. We also considered the standard peak signal-to-noise ratio (PSNR)~\cite{hore2010image},
$10\log_{10}\left(\mathrm{MAX}^2/\mathrm{MSE}\right)
$,
which compares the maximum possible pixel value, $\mathrm{MAX}$, with the mean-squared error (MSE), $(1/N)\sum_{i=1}^N (x_i - y_i)^2$. Note that PSNR is expressed on a log scale (dB), and higher values indicate lower reconstruction error.
The structural similarity index measure (SSIM)~\cite{hore2010image}, defined as
\begin{equation}
\mathrm{SSIM}(x,y)=
\frac{(2\mu_x\mu_y + C_1)(2\sigma_{xy}+C_2)}
{(\mu_x^2+\mu_y^2+C_1)(\sigma_x^2+\sigma_y^2+C_2)}\,,
\label{eq:ssim}
\end{equation}
evaluates similarity between reconstructed and groundtruth images in terms of luminance, contrast, and structural characteristics. In~\eqref{eq:ssim}, $\mu_x$ and $\mu_y$ represent mean intensities, $\sigma_x^2$ and $\sigma_y^2$ variances, and $\sigma_{xy}$ is the covariance between $x$ and $y$. The constants $C_1, C_2 > 0$ are used for numerical stability. The higher the SSIM, the more similar the images are, with the value of 1 indicating perfect similarity. We computed SSIM values separately for each spectral band and report the average across all the bands. 

To assess spectral reconstruction accuracy, we used spectral angle mapper (SAM)~\cite{yuhas1992discrimination} and erreur relative globale adimensionnelle de synthèse (ERGAS)~\cite{irmak2018map}. SAM is defined as
\begin{equation*}
\mathrm{SAM}(x,y)=
\cos^{-1}
\left(
\frac{
\sum_{i=1}^{C} x_i y_i
}{
\sqrt{\sum_{i=1}^{C} x_i^2}
\sqrt{\sum_{i=1}^{C} y_i^2}
}
\right)\,,
\end{equation*}
where $C$ denotes the number of spectral bands. 
It views each image as a vector in a high-dimensional space and measures their angular similarity, ignoring magnitude factors. 
In turn, ERGAS is defined as
\begin{equation*}
\mathrm{ERGAS}
=\frac{100}{r}
\sqrt{
\frac{1}{C}
\sum_{c=1}^{C}
\left(
\frac{\mathrm{RMSE}^{(c)}(x, y)}{\mu^{(c)}}
\right)^2\,,
}
\end{equation*}
where $r$ represents the ratio between the spatial resolutions of the high- and low-resolution images, taking the values 4 for $4\times$ upsampling and 16 for $16\times$ upsampling. Again, $C$ denotes the number of channels. The root mean squared error (RMSE) in the $c$th spectral band is defined as
$
\mathrm{RMSE}^{(c)}(x,y)
=
\sqrt{
(1/N_c)
\sum_{i=1}^{N_c}
\big(x_i^{(c)}-y_i^{(c)}\big)^2
}$,
where $x_i^{(c)}$ and $y_i^{(c)}$ denote the pixel values at location $i$ of the reconstructed and reference images in channel $c$, and $N_c$ is the total number of pixels in that band. Also, $\mu^{(c)} = (1/N_c) \sum_{i=1}^{N_c} y_i^{(c)}$ denotes the mean value of $y^{(c)}$. By normalizing the reconstruction error within each band, ERGAS allows comparing different methods in terms of their ability to preserve spectral information.

\modification{To statistically validate performance differences across SR models, we performed non-parametric Kruskal--Wallis tests followed by Dunn's post-hoc tests with Bonferroni correction~\cite{pohlert2014pairwise} for all metrics under each experimental setup. We first applied the Kruskal--Wallis test to assess whether at least one model differed from the others. The degrees of freedom for this test are $\text{df}=k-1=7$, where $k=8$ is the number of SR models compared. When the Kruskal--Wallis test indicated a significant difference, we performed pairwise comparisons between SR models using Dunn's post-hoc test. To reduce the risk of false positives due to multiple comparisons, we applied a Bonferroni correction.}



\subsection{\modification{Computational Complexity of Super-Resolution Methods}}
\label{sec:complexity}
\modification{
To assess the practical computational requirements of the evaluated methods, Table~\ref{tab:computational_complexity} reports the number of trainable parameters, floating-point operations (FLOPs), and measured inference time. Learning-based methods were evaluated using publicly released pretrained checkpoints without retraining or fine-tuning; therefore, training time is not applicable to the present study. FLOPs and inference time were measured for a 512$\times$256-pixel low-resolution RGN input at $4\times$ upscaling on an NVIDIA GeForce RTX 4090 Laptop GPU with 16\,GB memory. The reported inference times correspond to a single $4\times$ inference pass. Table~\ref{tab:computational_complexity} shows substantial differences in computational requirements among the evaluated SISR methods. Bicubic and nearest-neighbor interpolation required negligible processing time. Among the learning-based approaches, HiT-SR had the lowest parameter count (0.87\,M) and FLOPs (0.23\,T), with an inference time of 0.31\,s per image. Real-ESRGAN has the least  inferencing time (0.24\,s per image) even with more parameters and FLOPs than HiT-SR. HAT and SRFormer required 1.74\,s and 1.42\,s per image, respectively. The diffusion-based methods had the highest computational requirements. StableSR required 1554.64\,M parameters and 371.46\,T FLOPs, with a mean inference time of 32.42\,s per image. These results indicate that both reconstruction quality and inference cost should be considered when selecting an SR method for large-scale agricultural imaging.}

\begin{table*}[th]
\centering
\renewcommand{\arraystretch}{1.2}
\caption{\modification{Computational characteristics of the evaluated super‑resolution methods.
Learning‑based methods were evaluated using publicly released pretrained
checkpoints without retraining or fine‑tuning. FLOPs and inference time were
measured for an LR RGN input of 512$\times$256 pixels at 4$\times$ upscaling using an NVIDIA GeForce RTX 4090 (16\,GB, mobile) GPU.}}
\label{tab:computational_complexity}
\vspace{-0.2cm}
\begin{tabular}{l l l l} 
\toprule
\textbf{SR Technique} & \textbf{Parameters (M)} & \textbf{FLOPs (T)} & \textbf{Inference time (s/image)} \\ 
\midrule
Bicubic \cite{khaledyan2020low} & -- & -- & \phantom{0}$3.79 \times 10^{-4}$
 \\
Nearest neighbor \cite{fattal2007image} & -- & -- & \phantom{0}$3.39 \times 10^{-4}$ \\
HAT \cite{chen2025hat} & \phantom{00}20.77 & \phantom{00}5.48 & \phantom{0}1.74 \\
HiT‑SR \cite{zhang2024hit} & \phantom{000}0.87 & \phantom{00}0.23 & \phantom{0}0.31 \\
SRFormer \cite{zhou2023srformer} & \phantom{00}10.52 & \phantom{00}3.04 & \phantom{0}1.42 \\
Real‑ESRGAN \cite{wang2021real} & \phantom{00}16.70 & \phantom{00}4.70 & \phantom{0}0.24 \\
ResShift \cite{yue2023resshift, yue2024efficient} & \phantom{0}173.92 & \phantom{0}43.21 & \phantom{0}4.36 \\
Stable SR \cite{wang2024exploiting} & 1554.64 & 371.46 & 32.42 \\
\bottomrule
\end{tabular}

\vspace{0.1cm}
\noindent\footnotesize
\textit{Checkpoint details.} For reproducibility, the publicly released pretrained weights used for inference are: HAT:-\texttt{HAT\_SRx4\_ImageNet-pretrain.pth}; HiT-SR:-\texttt{HiT-SRF-4x.pth}; SRFormer:-\texttt{SRFormer\_SRx4\_DF2K.pth}; Real\-ESRGAN:-\texttt{RealESRGAN\_x4plus.pth}; ResShift--\texttt{resshift\_realsrx4\_s4\_v3.pth} with \texttt{autoencoder\_vq\_f4.pth}; StableSR--\texttt{stablesr\_turbo.ckpt} with \texttt{vqgan\_cfw\_00011.ckpt}. Bicubic and NN do not require pretrained weights.
\end{table*}

\vspace{0.1cm}

\subsection{Experiment 1: Whole-Image Reconstruction} \label{sec:ex1}

\begin{figure}[h]
\centering

\begin{subfigure}{0.9\linewidth}
    \centering
    \includegraphics[width=\linewidth]{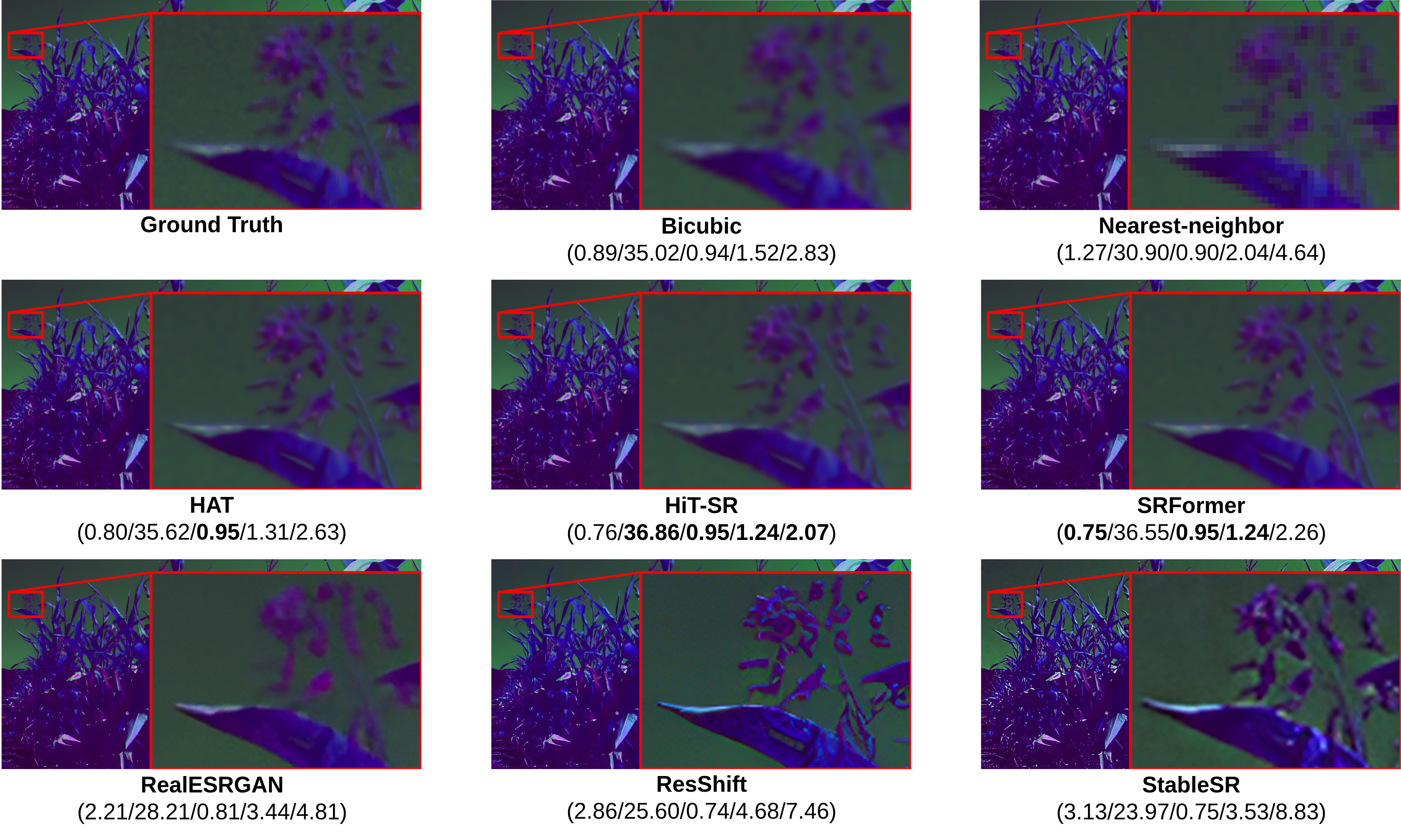}
    \caption{Maize}
    \label{fig:maize}
\end{subfigure}

\vspace{0.5em}

\begin{subfigure}{0.9\linewidth}
    \centering
    \includegraphics[width=\linewidth]{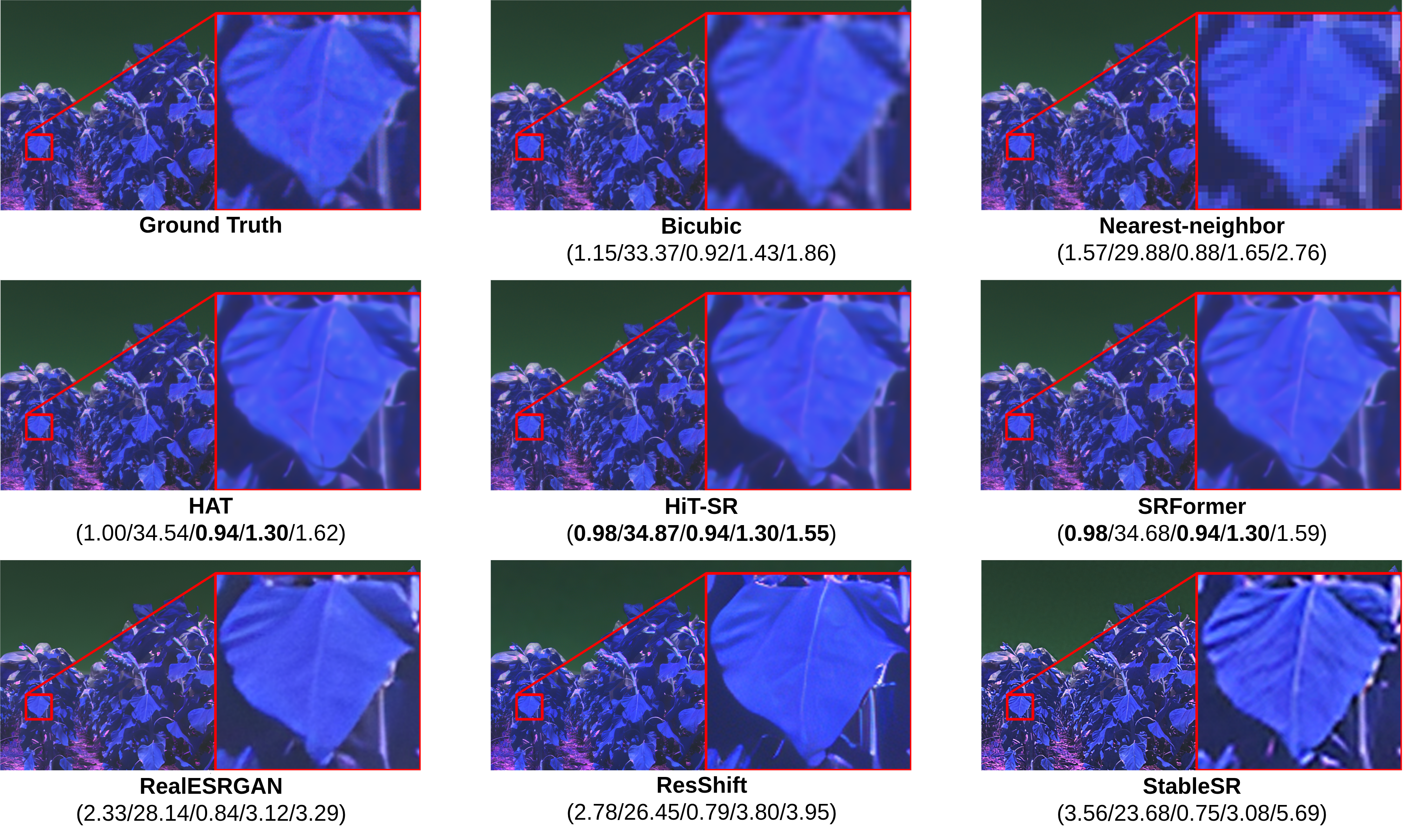}
    \caption{Sunflower}
    \label{fig:sunflower}
\end{subfigure}


\caption{\modification{Qualitative comparison of two representative $4\times$ single-image super-resolution (SISR) reconstruction examples for \text{(a)} maize and \text{(b)} sunflower crops. For each example, the full reconstructed image is shown together with an enlarged local region (zoomed-in view) to facilitate detailed comparison of crop structures, textures, and phenotypic characteristics. Results from all SISR methods are accompanied by the corresponding reconstruction metrics: MAE$\downarrow$, PSNR (dB)$\uparrow$, SSIM$\uparrow$, SAM ($^\circ$)$\downarrow$, and ERGAS$\downarrow$. The zoomed-in regions highlight differences in the preservation of fine-scale agricultural structures and spectral-spatial details. While the generative approaches (last row) produce visually sharper and more perceptually appealing reconstructions, they may introduce artificial structures or textures that are not supported by the groundtruth, indicating a potential trade-off between perceptual sharpness and faithful reconstruction of agriculturally relevant information.}}

\label{fig:single}
\end{figure}


\begin{figure}[h]
\centering

\begin{subfigure}{.90\linewidth}
    \centering
    \includegraphics[width=\linewidth]{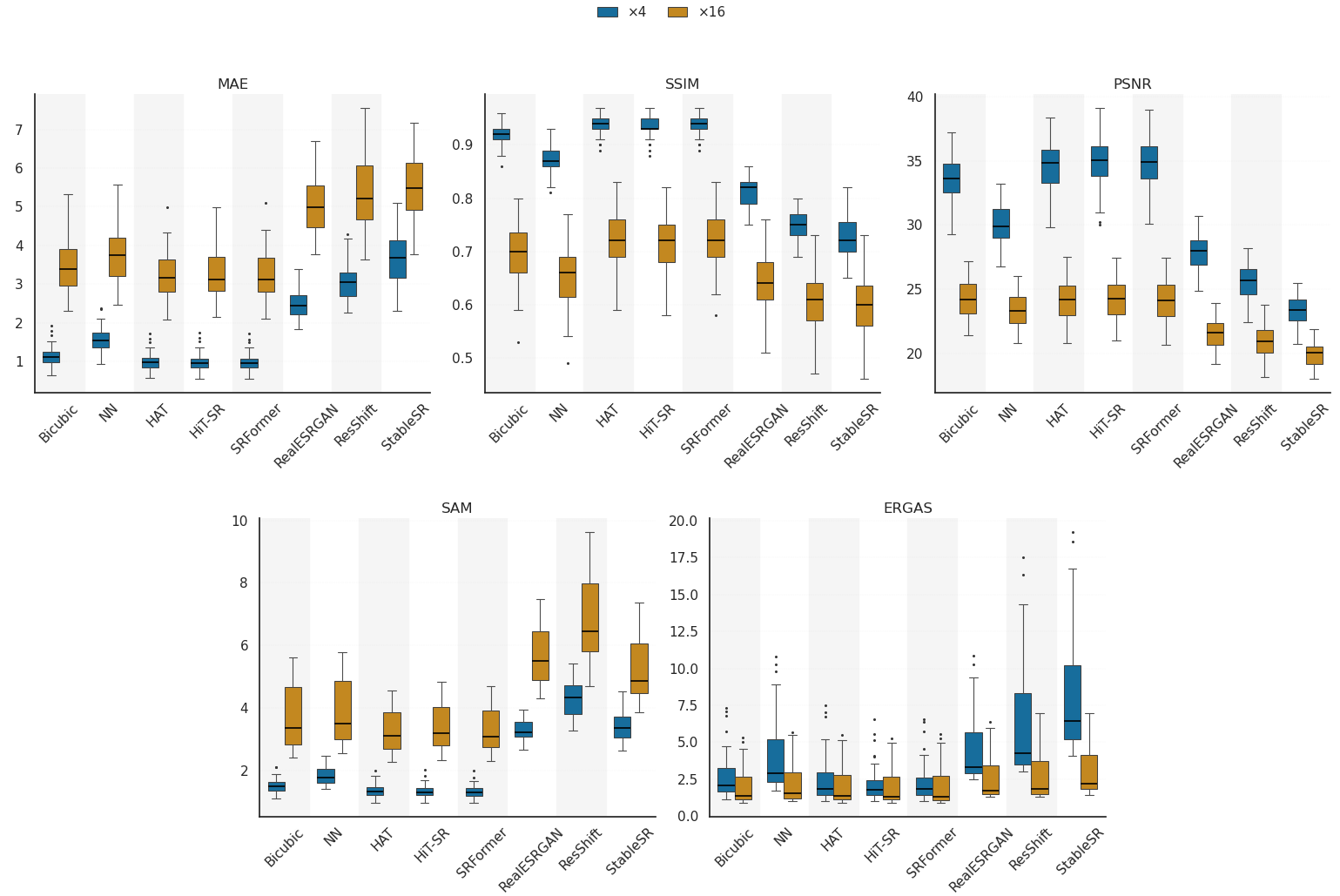}
    \caption{Whole image analysis}
    \label{fig:box_plot}
\end{subfigure}

\vspace{0.5em}

\begin{subfigure}{.90\linewidth}
    \centering
    \includegraphics[width=\linewidth]{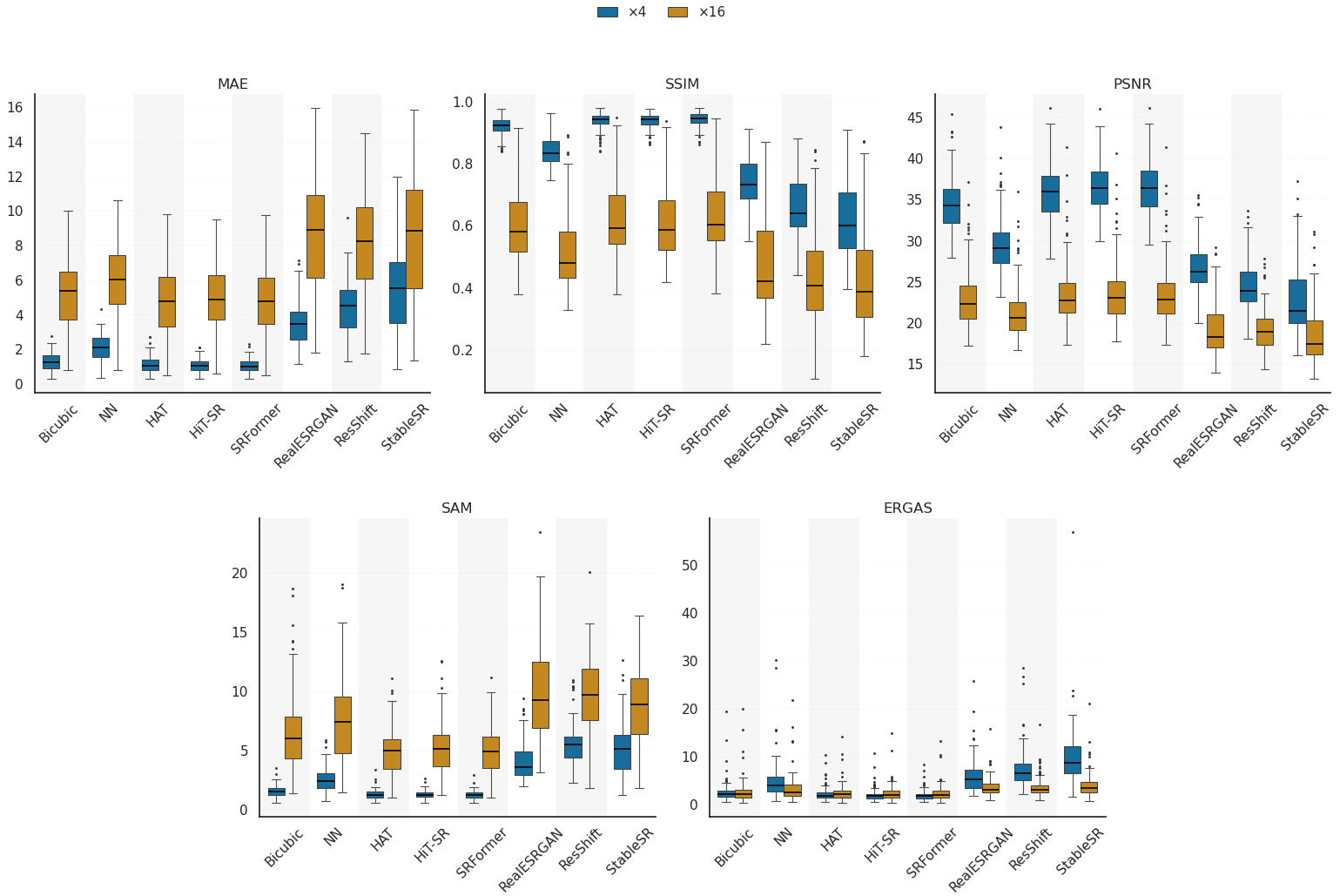}
    \caption{Masked region analysis}
    \label{fig:box_plot2}
\end{subfigure}

\caption{Distributional comparison of RGN image reconstruction quality across SR models at $4\times$ and $16\times$ scaling factors. Figure (a) shows the whole image results (Sec.~\ref{sec:ex1}) and (b) shows masked leaf region results (Sec.~\ref{sec:ex2}).}
\label{fig:box_main}
\end{figure}

Figure~\ref{fig:single} shows the reconstruction at a $4\times$ factor by all the methods of two representative whole-image RGN examples, one for a maize crop (Fig.~\ref{fig:maize}), another for a sunflower crop (Fig.~\ref{fig:sunflower}). In both examples, it can be observed that the transformer- and attention-based algorithms (HAT, HiT-SR, and SRFormer) obtained the best reconstruction performance, followed by the classical methods (bicubic and NN). The worst performance was obtained by generative methods (RealESRGAN, ResShift, and StableSR) which, despite producing visually-pleasing reconstructions, introduced artifacts inexistent in the groundtruth images. Figure~\ref{fig:box_plot} reports the results for each metric in the entire dataset for both scaling factors $4\times$ and $16\times$. The results are visualized using box plots, where the central horizontal line within each box denotes the median performance. The box boundaries represent the first and third quartiles (Q1–Q3), corresponding to the interquartile range (IQR). The whiskers extend to the most extreme data points within 1.5 times the IQR, while any samples outside this range are treated as outliers and plotted individually. \modification{To statistically validate performance differences given the skewed error distributions and outliers, we performed non-parametric Kruskal--Wallis tests followed by post-hoc Dunn's tests with Bonferroni correction. The omnibus tests confirmed significant differences among models across all metrics for both $4\times$ ($H(7) \in [324.14, 516.55]$, $p < 0.001$) and $16\times$ ($H(7) \in [76.00, 383.70]$, $p < 0.001$) setups.}

Fig.~\ref{fig:single} further validates this trend: for $4\times$ upsampling (blue boxes), transformer- and attention-based methods consistently achieve the strongest results. Specifically, SRFormer obtained the lowest MAE (0.98) and lowest SAM (1.32$^\circ$), while sharing the highest SSIM (0.94) with HAT (1.00 MAE, 1.34$^\circ$ SAM). HiT-SR achieved the highest PSNR (34.95~dB) and the lowest ERGAS (2.09), with only marginal differences relative to SRFormer and HAT. \modification{Post-hoc pairwise tests confirmed that all three Transformer models significantly outperformed RealESRGAN, ResShift, and StableSR across all metrics ($p < 0.001$, Bonferroni-adjusted), while differences among HAT, HiT-SR, and SRFormer were not statistically significant.} Despite its simplicity, bicubic interpolation remained a competitive baseline, achieving a PSNR of 33.58~dB and an SSIM of 0.92. In contrast, the generative approaches exhibited substantially larger reconstruction errors. For a $16\times$ factor (orange boxes), reconstruction quality deteriorated across all methods and exhibited larger variations. \modification{Transformer based approaches remained robust; HAT attained the lowest MAE (3.20), highest SSIM (0.73), and lowest SAM (3.24$^\circ$), closely followed by SRFormer (3.24 MAE, 0.72 SSIM, 3.29$^\circ$ SAM) and HiT-SR (3.26 MAE, 0.72 SSIM, 3.38$^\circ$ SAM), with no statistically significant differences within the Transformer family. Pairwise tests confirmed that Transformers maintained statistically superior MAE, PSNR, SSIM, and SAM over generative models ($p \le 0.008$)}. Bicubic interpolation attained the highest PSNR (24.27~dB) and matched HiT-SR's lowest ERGAS (1.98), while maintaining competitive fidelity relative to deep learning models at this extreme scaling factor. Generative and diffusion models continued to yield the lowest quantitative fidelity, with ResShift and StableSR recording the highest spectral distortion and reconstruction errors.

\modification{Table~\ref{tab:VI_whole} reports VI reconstruction results, with each column displaying the MAE, PSNR, and SSIM values for each VI index of Table~\ref{tab:paperVI}. Non-parametric Kruskal--Wallis tests confirmed statistically significant performance differences across models for all eight vegetation indices at both $4\times$ ($H(7) \in [395.95, 499.24]$, $p < 0.001$) and $16\times$ ($H(7) \in [76.00, 455.25]$, $p < 0.001$) scale factors. Post-hoc Dunn's tests with Bonferroni correction verified that Transformer architectures (HAT, HiT-SR, and SRFormer) significantly outperformed GAN- and diffusion-based baselines across all VI metrics ($p < 0.001$), while pairwise differences within the Transformer family were not statistically significant. It can be seen that HAT, SRFormer, and HiT-SR consistently achieve the strongest performance across both scaling factors, with HAT yielding the highest overall performance on most indices. At 4$\times$, SAVI and OSAVI were reconstructed most accurately, where HAT achieved 41.46 dB PSNR and 0.95 SSIM for SAVI (MAE of 1.78) and 39.41 dB PSNR and 0.93 SSIM for OSAVI (MAE of 1.92), with SRFormer and HiT-SR producing closely comparable results. The DVI and NDVI remain within a relatively high-quality reconstruction range, whereas ratio-based indices such as MSR and RVI are reconstructed with poorer accuracy across all evaluated methods. At 16$\times$, performance degradation becomes more noticeable, especially for the ratio-based indices. HAT maintained the highest accuracy across these indices, achieving 34.85 dB PSNR (MAE of 3.64, SSIM of 0.85) for SAVI and 32.90 dB PSNR (MAE of 3.86, SSIM of 0.80) for OSAVI, compared to 34.67 dB and 32.70 dB for SRFormer, and 34.52 dB and 32.54 dB for HiT-SR. In contrast, GAN and diffusion-based methods exhibited substantially lower reconstruction quality, with ResShift producing the lowest RVI reconstruction at 16$\times$, achieving an MAE of 64.72, a PSNR of 12.37 dB, and an SSIM of 0.44.}

\subsection{Experiment 2: Leaf-Level Reconstruction Analysis Within Masked Regions} \label{sec:ex2}

Figure~\ref{fig:box_plot2} and Table~\ref{tab:VI_leaf} show the results of leaf-level SR across 100 manually segmented leaf regions. Unlike Experiment~1, where reconstruction metrics were computed over the entire image, here metrics are computed exclusively in leaf regions. \modification{Non-parametric Kruskal--Wallis tests confirmed statistically significant performance variations across models across all raw bands and vegetation indices at both $4\times$ ($H(7) \in [338.31, 593.72]$, $p < 0.001$) and $16\times$ ($H(7) \in [101.41, 420.34]$, $p < 0.001$) scale factors.} Comparing Figs.~\ref{fig:box_plot} and~\ref{fig:box_plot2} for $4\times$,  we observe that reconstruction performance improves in terms of PSNR, SSIM, and ERGAS for the top-performing methods. SRFormer obtained the best metrics: MAE of 1.06, PSNR of 36.63~dB, SSIM of 0.94, SAM of 1.26$^\circ$, and ERGAS of 2.00. Relative to Experiment~1, PSNR increases from 34.86 to 36.63~dB for SRFormer and from 34.95 to 36.60~dB for HiT-SR, \modification{and from 34.64 to 36.12~dB for HAT}. Bicubic interpolation also shows improved performance, with PSNR increasing from 33.58~dB to 34.69~dB while maintaining an SSIM of 0.92. Diffusion-based and GAN-based methods obtained larger errors and lower spectral consistency for most metrics. At $16\times$, reconstruction performance decreases for all methods. SRFormer achieved the lowest MAE (4.73) and the highest SSIM (0.63), HAT obtained the lowest SAM (4.94$^\circ$), and HiT-SR achieved the highest PSNR (23.80~dB) alongside the lowest ERGAS (2.40) among the learning-based methods. 
\modification{Unlike Experiment~1 where bicubic interpolation achieved the top PSNR at $16\times$, here the transformer based models (HiT-SR at 23.80~dB, SRFormer at 23.76~dB, and HAT at 23.55~dB) outperform bicubic interpolation (23.16~dB).} Compared with Experiment~1, all methods exhibit reduced performance at $16\times$, and differences between learning-based methods and bicubic interpolation are less pronounced under this scaling factor. As shown in Fig.~\ref{fig:box_plot2}, generative models exhibit higher variability and more outliers.

\FloatBarrier
\begin{sidewaystable}[t]
\scriptsize
\caption{Quantitative evaluation of vegetation index (VI) reconstruction derived from RGN imagery for different SR methods at $4\times$ and $16\times$ scaling factors using MAE $\downarrow$, PSNR (dB) $\uparrow$, and SSIM $\uparrow$. Best results in \textbf{bold}, and worst \underline{underlined}. For readability, MAE values are scaled by 100. \text{(a)} Whole-image experiments. \text{(b)} Leaf-level experiments.}
\renewcommand{\arraystretch}{1.1}
\setlength{\tabcolsep}{3pt}

\vspace{0.3em}
\scalebox{0.95}{
\begin{subtable}[t]{0.95\textwidth}
\centering
\caption{Performance of super-resolution methods evaluated across the entire image.}
\label{tab:VI_whole}
\begin{NiceTabular}{@{}l r c c c c c c c c@{}}
\CodeBefore
\rowcolors{3}{black!10}{}[respect-blocks]
\Body
\toprule
\multirow{2}{*}{\textbf{SR Technique}} 
& \multirow{2}{*}{\textbf{Scale}} & \textbf{DVI} & \textbf{GNDVI} & \textbf{MSAVI} & \textbf{MSR} & \textbf{NDVI} & \textbf{OSAVI} & \textbf{RVI} &\textbf{SAVI} \\
\cmidrule(lr){3-3}
\cmidrule(lr){4-4}
\cmidrule(lr){5-5}
\cmidrule(lr){6-6}
\cmidrule(lr){7-7}
\cmidrule(lr){8-8}
\cmidrule(lr){9-9}
\cmidrule(lr){10-10}
& & MAE / PSNR / SSIM & MAE / PSNR / SSIM & MAE / PSNR / SSIM 
& MAE / PSNR / SSIM & MAE / PSNR / SSIM & MAE / PSNR / SSIM & MAE / PSNR / SSIM & MAE / PSNR / SSIM \\
\midrule
\arrayrulecolor{black!10}

\Block{2-1}{Bicubic}
& $4\times$
 
& $\phantom{0}$1.69 / 37.66 / 0.92  
& $\phantom{0}$2.15 / 34.57 / 0.91  
& $\phantom{0}$1.88 / 37.34 / 0.90  
& $\phantom{0}$4.27 / 32.68 / 0.82  
& $\phantom{0}$2.18 / 36.42 / 0.88  
& $\phantom{0}$2.04 / 38.80 / 0.91  
& $\phantom{0}$8.47 / 34.07 / 0.86  
& $\phantom{0}$1.91 / 40.80 / 0.94\\

& $16\times$

& $\phantom{0}$3.59 / 30.94 / 0.78
& $\phantom{0}$6.92 / 26.19 / 0.70
& $\phantom{0}$3.79 / 31.07 / 0.76
& $\phantom{0}$8.48 / \textbf{26.45} / 0.59
& $\phantom{0}$4.23 / \textbf{30.51} / \textbf{0.72}
& $\phantom{0}$4.01 / 32.77 / 0.79
& 17.13 / \textbf{27.65} / 0.69
& $\phantom{0}$3.83 / 34.60 / 0.84\\

\midrule[0.1pt]

\Block{2-1}{Nearest Neighbor}
& $4\times$

& $\phantom{0}$2.03 / 35.63 / 0.89
& $\phantom{0}$2.79 / 32.00 / 0.87
& $\phantom{0}$2.20 / 35.60 / 0.87
& $\phantom{0}$4.92 / 31.10 / 0.78
& $\phantom{0}$2.49 / 34.94 / 0.85
& $\phantom{0}$2.36 / 37.23 / 0.89
& $\phantom{0}$9.83 / 32.44 / 0.83
& $\phantom{0}$2.23 / 39.11 / 0.92\\

& $16\times$

& $\phantom{0}$3.86 / 30.12 / 0.75
& $\phantom{0}$7.03 / 25.55 / 0.67
& $\phantom{0}$4.03 / 30.32 / 0.73
& $\phantom{0}$8.92 / 25.79 / 0.56
& $\phantom{0}$4.42 / 29.92 / 0.69
& $\phantom{0}$4.23 / 32.12 / 0.77
& 18.10 / 26.98 / 0.66
& $\phantom{0}$4.06 / 33.88 / 0.82\\

\midrule[0.1pt]
\Block{2-1}{HAT}
& $4\times$

& $\phantom{0}$\textbf{1.55} / \textbf{38.45 }/ \textbf{0.93}
& $\phantom{0}$1.82	/ 36.04 / 0.93
& $\phantom{0}$\textbf{1.75} / \textbf{38.01} / 0.92
& $\phantom{0}$\textbf{4.01} / \textbf{33.17} / \textbf{0.84}
& $\phantom{0}$\textbf{2.05} / \textbf{37.00} / \textbf{0.89}
& $\phantom{0}$\textbf{1.92} / \textbf{39.41} / \textbf{0.93}
& $\phantom{0}$\textbf{7.97} / 33.95 / \textbf{0.88}
& $\phantom{0}$\textbf{1.78} / \textbf{41.46} / \textbf{0.95}\\
& $16\times$

& $\phantom{0}$\textbf{3.35} / \textbf{31.36} / \textbf{0.79}
& $\phantom{0}$\textbf{5.36} / \textbf{26.94} / \textbf{0.74}
& $\phantom{0}$\textbf{3.60} / \textbf{31.34} / \textbf{0.77}
& $\phantom{0}$\textbf{8.33 }/ 26.18 / \textbf{0.60}
& $\phantom{0}$\textbf{4.11} / 30.47 /\textbf{ 0.72}
& $\phantom{0}$\textbf{3.86	}/ \textbf{32.90 }/\textbf{ 0.80}
& \textbf{17.06} / 26.67 /\textbf{ 0.70}
& $\phantom{0}$\textbf{3.64} / \textbf{34.85} / \textbf{0.85}\\

\midrule[0.1pt]

\Block{2-1}{HiT-SR}
& $4\times$

& $\phantom{0}$1.57 / 38.39 / \textbf{0.93}
& $\phantom{0}$1.80 / 36.41 / 0.93
& $\phantom{0}$1.77 / 37.93 / 0.91
& $\phantom{0}$4.05 / 33.11 / \textbf{0.84}
& $\phantom{0}$2.07 / 36.91 / \textbf{0.89}
& $\phantom{0}$1.94 / 39.32 / 0.92
& $\phantom{0}$8.06 / \textbf{34.11} / \textbf{0.88}
& $\phantom{0}$1.80 / 41.37 / \textbf{0.95}\\

& $16\times$

& $\phantom{0}$3.45 / 31.16 / 0.78
& $\phantom{0}$5.72 / 26.87 / 0.73
& $\phantom{0}$3.75 / 31.01 / 0.76
& $\phantom{0}$8.74 / 25.84 / 0.59
& $\phantom{0}$4.29 / 30.14 / 0.71
& $\phantom{0}$4.03 / 32.54 / 0.79
& 17.98 / 26.47 / 0.68
& $\phantom{0}$3.79 / 34.52 / 0.84\\

\midrule[0.1pt]

\Block{2-1}{SRFormer}
& $4\times$

& $\phantom{0}$1.56 / 38.44 / \textbf{0.93}
& $\phantom{0}$\textbf{1.72 }/ \textbf{36.49} / \textbf{0.94}
& $\phantom{0}$1.76 / 37.97 / \textbf{0.92}
& $\phantom{0}$4.04 / 33.12 / \textbf{0.84}
& $\phantom{0}$2.06 / 36.95 / \textbf{0.89}
& $\phantom{0}$1.93 / 39.36 / \textbf{0.93}
& $\phantom{0}$8.03 / 33.94 / \textbf{0.88}
& $\phantom{0}$1.79 / 41.42 / \textbf{0.95}\\

& $16\times$

& $\phantom{0}$3.41 / 31.25 / \textbf{0.79}
& $\phantom{0}$5.39 / 26.88 / \textbf{0.74}
& $\phantom{0}$3.68 / 31.16 / 0.76
& $\phantom{0}$8.53 / 26.00 / 0.59
& $\phantom{0}$4.20 / 30.29 / \textbf{0.72}
& $\phantom{0}$3.95/ 32.70 / 0.79
& 17.49 / 26.55 / 0.69
& $\phantom{0}$3.72 / 34.67 / 0.84\\

\midrule[0.1pt]

\Block{2-1}{Real ESRGAN}
& $4\times$

& $\phantom{0}$3.35 / 32.26 / 0.82
& $\phantom{0}$4.87 / 28.50 / 0.79
& $\phantom{0}$3.83 / 31.58 / 0.79
& $\phantom{0}$9.09 / 26.24 / 0.63
& $\phantom{0}$4.53 / 30.35 / 0.74
& $\phantom{0}$4.21 / 32.87 / 0.81
& 18.40 / 26.80 / 0.72
& $\phantom{0}$3.89 / 35.03 / 0.86\\

& $16\times$

& $\phantom{0}$5.58 / 27.70 / 0.70
& $\phantom{0}$9.16 / 23.26 / 0.64
& $\phantom{0}$6.19 / 27.39 / 0.68
& 14.41 / 22.21 / 0.50
& $\phantom{0}$7.14 / 26.46 / 0.64
& $\phantom{0}$6.69 / 28.90 / 0.72
& 29.33 / 22.65 / 0.64
& $\phantom{0}$6.26 / 30.90 / 0.77\\

\midrule[0.1pt]

\Block{2-1}{ResShift}
& $4\times$

& $\phantom{0}$\underline{5.19} / \underline{28.57} / \underline{0.70}
& $\phantom{0}$\underline{5.18} / 26.87 / \underline{0.74}
& $\phantom{0}$\underline{5.93} / \underline{27.94} / \underline{0.65}
& \underline{15.09} / \underline{20.89} / \underline{0.43}
& $\phantom{0}$\underline{6.92} / \underline{26.86} / \underline{0.59}
& $\phantom{0}$\underline{6.47} / \underline{29.33} / \underline{0.68}
& \underline{34.52} / \underline{17.2} / \underline{0.54}
& $\phantom{0}$\underline{6.01} / \underline{31.44} / \underline{0.75}\\

& $16\times$

& $\phantom{0}$\underline{8.86} / \underline{24.26} / \underline{0.57}
& $\phantom{0}$\underline{8.88} / \underline{22.48} / \underline{0.61}
& $\phantom{0}$\underline{9.86} / \underline{23.87} / \underline{0.54}
& \underline{25.62} / \underline{16.52} / \underline{0.32}
& \underline{11.04} / \underline{23.16} / \underline{0.50}
& \underline{10.47} / \underline{25.51} / \underline{0.59}
& \underline{64.72} / \underline{12.37} / \underline{0.44}
& $\phantom{0}$\underline{9.89} / \underline{27.45} / \underline{0.65}\\

\midrule[0.1pt]

\Block{2-1}{Stable SR}
& $4\times$

& $\phantom{0}$3.65 / 30.78 / 0.78
& $\phantom{0}$5.09 / \underline{26.67} / \underline{0.74}
& $\phantom{0}$3.96 / 30.63 / 0.75
& $\phantom{0}$9.82 / 24.02 / 0.57
& $\phantom{0}$4.71 / 29.24 / 0.69
& $\phantom{0}$4.33 / 31.99 / 0.77
& 20.95 / 22.27 / 0.66
& $\phantom{0}$4.02 / 34.13 / 0.83\\

& $16\times$

& $\phantom{0}$6.07 / 26.08 / 0.67
& $\phantom{0}$8.23 / 23.22 / 0.63
& $\phantom{0}$6.30 / 26.37 / 0.65
& 14.51 / 20.86 / 0.47
& $\phantom{0}$\underline{6.92} / 25.83 / 0.60
& $\phantom{0}$6.60 / 28.14 / 0.68
& 31.19 / 19.67 / 0.59
& $\phantom{0}$6.34 / 29.92 / 0.74\\

\arrayrulecolor{black}
\bottomrule[1pt]
\end{NiceTabular}
\end{subtable}
}
\vspace{1.2em}

\scalebox{.95}{
\begin{subtable}[t]{0.95\textwidth}
\centering
\caption{Performance of super-resolution methods evaluated within manually segmented leaf regions.}
\label{tab:VI_leaf}
\begin{NiceTabular}{@{}l r c c c c c c c c@{}}
\CodeBefore
\rowcolors{3}{black!10}{}[respect-blocks]
\Body
\toprule
\multirow{2}{*}{\textbf{SR Technique}}
& \multirow{2}{*}{\textbf{Scale}} & \textbf{DVI} & \textbf{GNDVI} & \textbf{MSAVI} & \textbf{MSR} & \textbf{NDVI} & \textbf{OSAVI} & \textbf{RVI} &\textbf{SAVI} \\
\cmidrule(lr){3-3}
\cmidrule(lr){4-4}
\cmidrule(lr){5-5}
\cmidrule(lr){6-6}
\cmidrule(lr){7-7}
\cmidrule(lr){8-8}
\cmidrule(lr){9-9}
\cmidrule(lr){10-10}
& & MAE / PSNR / SSIM & MAE / PSNR / SSIM & MAE / PSNR / SSIM 
& MAE / PSNR / SSIM & MAE / PSNR / SSIM & MAE / PSNR / SSIM & MAE / PSNR / SSIM & MAE / PSNR / SSIM \\
\midrule
\arrayrulecolor{black!10}

\Block{2-1}{Bicubic}
& $4\times$

& $\phantom{0}$1.83 / 38.58 / 0.93
& $\phantom{0}$2.90 / 33.09 / 0.92
& $\phantom{0}$1.73 / 39.04 / 0.93
& $\phantom{0}$4.60 / 27.57 / 0.79
& $\phantom{0}$1.71 / 39.05 / 0.93
& $\phantom{0}$1.70 / 41.08 / 0.95
& 11.60 / 27.30 / 0.79
& $\phantom{0}$1.71 / 42.67 / 0.96\\

& $16\times$

& $\phantom{0}$6.34 / 28.59 / 0.71
& 14.46 / 21.07 / 0.53
& $\phantom{0}$5.53 / 29.72 / 0.74
& 13.46 / \textbf{18.85} / 0.37
& $\phantom{0}$5.06 / \textbf{30.34} / \textbf{0.75}
& $\phantom{0}$5.20 / 32.11 / \textbf{0.80}
& 33.66 / 18.65 / 0.37
& $\phantom{0}$5.43 / 33.39 / 0.83\\

\midrule[0.1pt]

\Block{2-1}{Nearest Neighbor}
& $4\times$

& $\phantom{0}$2.73 / 34.64 / 0.87
& $\phantom{0}$5.02 / 28.18 / 0.83
& $\phantom{0}$2.51 / 35.36 / 0.88
& $\phantom{0}$6.32 / 24.51 / 0.70
& $\phantom{0}$2.39 / 35.75 / 0.88
& $\phantom{0}$2.41 / 37.62 / 0.90
& 15.78 / 24.44 / 0.70
& $\phantom{0}$2.47 / 39.04 / 0.92\\

& $16\times$

& $\phantom{0}$7.34 / 26.99 / 0.62
& 16.34 / 19.77 / 0.43
& $\phantom{0}$6.48 / 27.95 / 0.65
& 15.38 / 17.41 / 0.27
& $\phantom{0}$5.90 / 28.62 / 0.67
& $\phantom{0}$6.08 / 30.37 / 0.73
& 37.93 / 17.43 / 0.28
& $\phantom{0}$6.35 / 31.65 / 0.77
\\

\midrule[0.1pt]
\Block{2-1}{HAT}
& $4\times$

& $\phantom{0}$1.59 / 39.86 / \textbf{0.95}
& $\phantom{0}$2.41 / 35.43 / \textbf{0.95}
& $\phantom{0}$1.56 / 39.99 / \textbf{0.95}
& $\phantom{0}$4.29 / 28.22 / \textbf{0.82}
& $\phantom{0}$1.59 / 39.77 / \textbf{0.94}
& $\phantom{0}$1.56 / 41.90 / \textbf{0.96}
& 10.88 / 27.91 / \textbf{0.82}
& $\phantom{0}$1.55 / 43.62 / \textbf{0.97}\\
& $16\times$

& $\phantom{0}$\textbf{5.30} / \textbf{29.80} / \textbf{0.76}
& 9.99 / 23.50 / \textbf{0.63}
& $\phantom{0}$\textbf{4.78} / \textbf{30.57} / \textbf{0.76}
& \textbf{13.25} / 18.43 / \textbf{0.41}
& $\phantom{0}$\textbf{4.80} / 30.32 / 0.74
& $\phantom{0}$\textbf{4.70} / \textbf{32.57} / \textbf{0.80}
& \textbf{34.64} / 17.71 / \textbf{0.41}
& $\phantom{0}$\textbf{4.71} / \textbf{34.24} / \textbf{0.85}\\

\midrule[0.1pt]

\Block{2-1}{HiT-SR}
& $4\times$

& $\phantom{0}$\textbf{1.54} / \textbf{40.07} / \textbf{0.95}
& $\phantom{0}$2.32 / 35.73 / 0.95
& $\phantom{0}$\textbf{1.54} / 40.04 / \textbf{0.95}
& $\phantom{0}$4.30 / 28.13 / 0.81
& $\phantom{0}$1.59 / 39.70 / \textbf{0.94}
& $\phantom{0}$1.55 / 41.88 / \textbf{0.96}
& 10.90 / 27.81 / 0.81
& $\phantom{0}$\textbf{1.52} / 43.66 / \textbf{0.97}\\

& $16\times$

& $\phantom{0}$5.49 / 29.59 / 0.74
& 10.71 / 23.40 / 0.62
& $\phantom{0}$5.13 / 30.07 / 0.74
& 14.48 / 17.87 / 0.38
& $\phantom{0}$5.20 / 29.81 / 0.72
& $\phantom{0}$5.07 / 32.04 / 0.79
& 37.92 / 17.12 / 0.38
& $\phantom{0}$5.04 / 33.75 / 0.83\\
\midrule[0.1pt]

\Block{2-1}{SRFormer}
& $4\times$

& $\phantom{0}$\textbf{1.54} / 40.06 / \textbf{0.95}
& $\phantom{0}$\textbf{2.16} / \textbf{36.31} / \textbf{0.95}
& $\phantom{0}$\textbf{1.54} / \textbf{40.10} / \textbf{0.95}
& $\phantom{0}$\textbf{4.25} / \textbf{28.26} / \textbf{0.82}
& $\phantom{0}$\textbf{1.58} / \textbf{39.82} / \textbf{0.94}
& $\phantom{0}$\textbf{1.54} / \textbf{41.97} / \textbf{0.96}
& \textbf{10.77} / \textbf{27.95} / \textbf{0.82}
& $\phantom{0}$\textbf{1.52} / \textbf{43.72} / \textbf{0.97}\\

& $16\times$

& $\phantom{0}$5.39 / 29.71 / 0.75
& $\phantom{0}$\textbf{9.85} / \textbf{23.67} / \textbf{0.63}
& $\phantom{0}$4.87 / 30.42 / \textbf{0.76}
& 13.31 / 18.48 / 0.40
& $\phantom{0}$4.83 / 30.31 / 0.74
& $\phantom{0}$4.75 / 32.48 / \textbf{0.80}
& \textbf{34.64} / \textbf{17.85} / 0.40
& $\phantom{0}$4.79 / 34.11 / 0.84\\

\midrule[0.1pt]

\Block{2-1}{Real ESRGAN}
& $4\times$

& $\phantom{0}$4.61 / 30.93 / 0.83
& $\phantom{0}$6.71 / 26.69 / 0.79
& $\phantom{0}$4.77 / 30.70 / 0.81
& 13.78 / 18.45 / 0.45
& $\phantom{0}$5.17 / 30.06 / 0.77
& $\phantom{0}$4.94 / 32.38 / 0.83
& 34.78 / 17.99 / 0.45
& $\phantom{0}$4.73 / 34.31 / 0.87\\

& $16\times$

& 11.05 / 23.87 / 0.61
& 16.96 / 19.41 / 0.45
& 11.17 / 23.90 / 0.62
& 29.01 / 12.59 / 0.21
& 11.72 / 23.69 / 0.61
& 11.38 / 25.80 / 0.68
& 69.34 / 12.49 / 0.22
& 11.09 / 27.52 / 0.73\\

\midrule[0.1pt]

\Block{2-1}{ResShift}
& $4\times$

& $\phantom{0}$\underline{7.87} / \underline{26.42} / \underline{0.69}
& $\phantom{0}$8.42 / 25.24 / 0.73
& $\phantom{0}$\underline{8.04} / \underline{26.23} / \underline{0.66}
& \underline{26.67} / \underline{11.96} / \underline{0.30}
& $\phantom{0}$\underline{8.23} / \underline{26.03} / \underline{0.64}
& $\phantom{0}$\underline{8.06} / \underline{28.16} / \underline{0.71}
& \underline{82.00} /  \underline{8.32} / \underline{0.28}
& $\phantom{0}$\underline{7.91} / \underline{29.91} / \underline{0.76}\\

& $16\times$

& \underline{12.62} / \underline{22.45} / \underline{0.50}
& 15.71 / 20.29 / 0.46
& \underline{12.75} / \underline{22.36} / \underline{0.49}
& \underline{47.41} /  \underline{6.75} / \underline{0.12}
& \underline{12.80} / \underline{22.37} / \underline{0.48}
& \underline{12.55} / \underline{24.48} / \underline{0.57}
& \underline{183.33} /  \underline{0.82} / \underline{0.11}
& \underline{12.40} / \underline{26.15} / \underline{0.63}
\\

\midrule[0.1pt]

\Block{2-1}{Stable SR}
& $4\times$
& $\phantom{0}$6.23 / 28.46 / 0.74
& $\phantom{0}$\underline{9.43} / \underline{23.76} / \underline{0.67}
& $\phantom{0}$5.79 / 29.03 / 0.72
& 17.65 / 16.28 / 0.37
& $\phantom{0}$6.02 / 28.65 / 0.68
& $\phantom{0}$5.80 / 30.94 / 0.76
& 49.13 / 14.94 / 0.37
& $\phantom{0}$5.71 / 32.69 / 0.81\\

& $16\times$
& 11.88 / 23.03 / 0.53
& \underline{16.02} / \underline{19.37} / \underline{0.42}
& 10.75 / 23.77 / 0.54
& 29.34 / 11.90 / 0.19
& 10.37 / 23.85 / 0.54
& 10.43 / 25.87 / 0.61
& 82.94 / 10.35 / 0.20
& 10.65 / 27.38 / 0.66
\\

\arrayrulecolor{black}
\bottomrule[1pt]
\end{NiceTabular}
\end{subtable}}
\end{sidewaystable}
\FloatBarrier

The VI reconstruction results in Table~\ref{tab:VI_leaf} follow a similar pattern. \modification{Non-parametric Kruskal-Wallis tests confirmed statistically significant performance differences across models for all eight vegetation indices at both $4\times$ ($H(7) \in [338.31, 593.72]$, $p < 0.001$) and $16\times$ ($H(7) \in [101.41, 420.34]$, $p < 0.001$) scale factors. Post-hoc Dunn's tests with Bonferroni correction verified that Transformer architectures (HAT, HiT-SR, and SRFormer) significantly outperformed GAN- and diffusion-based baselines across all VI metrics ($p < 0.001$), while pairwise differences within the Transformer family were not statistically significant.} At $4\times$, soil-adjusted indices (SAVI and OSAVI) were the best reconstructed indices within leaf regions. SRFormer achieved a PSNR of 43.72~dB (resp.\ 41.97~dB) and an SSIM of 0.97 (resp.\ 0.96) for SAVI (resp.\ OSAVI).  HiT-SR and HAT produce comparable results for both indices, with HiT-SR achieving 43.66~dB (SAVI) and 41.88~dB (OSAVI), and HAT achieving 43.62~dB (SAVI) and 41.90~dB (OSAVI). In contrast, ratio-based indices such as MSR and RVI exhibited larger reconstruction errors and lower structural similarity for all the methods. At $16\times$, reconstruction quality decreased for all vegetation indices. HAT achieved the highest SAVI performance, with an MAE of 4.71, a PSNR of 34.24~dB, and an SSIM of 0.85. HAT also achieved the best OSAVI results, with an MAE of 4.70, a PSNR of 32.57~dB, and an SSIM of 0.80. SRFormer and HiT-SR produced similar, but slightly lower, results for these two indices. Reconstruction was weaker for ratio-based indices, most notably RVI. ResShift had the lowest RVI reconstruction quality at $16\times$, with an MAE of 183.33, a PSNR of 0.82~dB, and an SSIM of 0.11.

\section{Discussion}

The experimental results reveal key insights about applying RGB-pretrained SISR models to multispectral crop images in terms of spatial reconstruction quality and spectral consistency.

\begin{enumerate}

\item Behavior of super-resolution methods in multispectral reconstruction. \modification{Non-parametric Kruskal--Wallis tests followed by Dunn's post-hoc tests with Bonferroni correction confirmed that} Transformer-based methods (SRFormer, HiT-SR, and HAT) significantly outperform GAN and diffusion-based approaches across both whole-image and leaf-level evaluations. This improvement is most evident in spectral metrics (SAM and ERGAS), indicating better preservation of inter-band relationships in RGN images compared to generative models. Although SRFormer, HiT-SR, and HAT achieve comparable performance overall, SRFormer shows slightly more stable behavior across scaling factors. \modification{Pairwise comparisons revealed no statistically significant differences among SRFormer, HiT-SR, and HAT, placing all three transformer architectures within the same highest-performing statistical category.} In contrast, GAN-based (RealESRGAN) and diffusion-based models (ResShift, StableSR) produce visually sharper reconstructions but exhibit higher spectral distortion. This reflects their bias toward perceptual realism rather than radiometric accuracy. Bicubic interpolation remains a strong baseline in several cases and achieves competitive PSNR at high scaling factors compared to more complex pretrained models. It thus remains a robust, simple reference method for RGN-image reconstruction.

\item Impact of the upscaling factor on the quality of reconstruction. The scaling factor has a substantial impact on all the evaluated methods. \modification{At 4$\times$, transformer-based models (SRFormer, HiT-SR, and HAT) significantly outperform classical interpolation methods (bicubic and nearest-neighbor) across both whole-image and leaf-level evaluations ($p < 0.001$). However, at 16$\times$, the performance gap narrows considerably, with bicubic interpolation achieving statistically comparable PSNR and ERGAS values to Transformer models in several cases.} This reflects the severe loss of fine image detail, making accurate reconstruction difficult regardless of model type. The performance gap between learned SISR methods and bicubic interpolation becomes much smaller in terms of PSNR and SSIM. This indicates that when the input resolution is highly limited, there is insufficient information to reliably reconstruct fine details and, consequently, reconstruction is likely driven by simple interpolation behavior rather than learned model priors. As a result, increasing model complexity provides limited additional benefit, highlighting a fundamental limitation of SISR for multispectral images under extreme downsampling. We emphasize that the evaluated SISR models were originally designed for RGB images, which may further influence their performance when applied to multispectral data without fine-tuning.

\item Spectral consistency and vegetation index stability.
The results highlight that spectral preservation is an important aspect of multispectral SR for vegetation analysis. Transformer-based approaches consistently achieved lower SAM and ERGAS values, indicating stronger preservation of spectral relationships and radiometric consistency. This advantage translated directly to VI reconstruction, where transformer-based methods obtained better performance across most indices. However, reconstruction reliability also depends on the VI formulation. Soil-adjusted indices such as SAVI and OSAVI remained comparatively robust, while ratio-based indices such as RVI and MSR were considerably more sensitive to reconstruction errors, especially at $16\times$. Since ratio formulations amplify small inter-band deviations, even minor spectral inaccuracies propagated into disproportionately larger index errors. This sensitivity persisted even in leaf-level evaluation, suggesting that the observed instability originates primarily from spectral error amplification rather than heterogeneous backgrounds. These findings confirm that visually convincing SR outputs alone are insufficient for quantitative agricultural analysis, especially when spectral relationships are not preserved.

\item Generalization to multispectral imagery and methodological limitations.
Although the evaluated SR methods were originally developed for RGB imagery, we directly applied them to RGN multispectral data without domain-specific fine-tuning. \modification{The key challenge is to enhance spatial detail while preserving spectral information and reliable vegetation-index analysis for crop characterization and phenotyping. The evaluated methods showed varying performance across architectures and evaluation conditions, with transformer-based architectures exhibiting the strongest overall performance.} We acknowledge some limitations of our study. The evaluated methods were used directly without multispectral fine-tuning, which likely constrained the reconstruction performance, especially under extreme scaling factors. We also used RGN images acquired with a single acquisition configuration. Considering additional spectral bands, sensor types, or crop conditions could have led to different results. \modification{In addition, our imagery was acquired at ground level using a horizontal or near-horizontal viewing perspective, whereas multispectral remote sensing imagery is commonly acquired from a nadir-view perspective. Due to equipment limitations, we were unable to acquire comparable nadir-view imagery, limiting our ability to evaluate the performance of SR methods under different viewing perspectives. Differences in perspective, depth-dependent scale, leaf orientation, and occlusion may influence the spatial characteristics of the imagery and the generalization of pretrained SR models.}
Although leaf-level evaluation provided biologically meaningful analysis, manual segmentation is not scalable and may introduce annotation uncertainty. Since leaf morphology is irregular and masked regions are not perfectly rectangular, residual background pixels may remain during annotation, masking, or reconstruction padding, introducing minor background interference into leaf-level evaluation. \modification{These limitations may affect the reliability and generalizability of SR-based spatial and spectral measurements for crop characterization and phenotyping. Future work will compare SISR models developed for multispectral and hyperspectral imagery with the RGB-based models evaluated in this study. These findings highlight the need for future approaches that better preserve spectral relationships while enhancing spatial detail, with the goal of improving the reliability of super-resolved imagery for crop characterization and phenotyping. Since the current imagery was acquired from a horizontal or near-horizontal viewing perspective, further investigation is needed to determine how viewing geometry affects reconstruction quality, crop characterization, and phenotyping outcomes compared with the nadir perspective commonly used in remote sensing. The analysis could also be extended beyond the RGN bands to assess whether model performance is consistent across spectral bands with different spectral responses. Testing additional sensors, viewing perspectives, crop types, and growth stages would provide a broader assessment of model robustness and the practical benefits and limitations of super-resolution for crop characterization and phenotyping.}

\item Implications for precision agriculture applications. The results demonstrate that visual reconstruction quality is a poor guide to evaluate SR methods in agricultural remote sensing. For downstream phenotyping tasks, spectral consistency plays a role more important than perceptual realism. Transformer-based approaches provide the most balanced trade-off between spatial enhancement and spectral preservation, making them the most suitable among the evaluated methods for crop monitoring applications. However, at $16\times$, the observed degradation highlights a practical limitation, as extreme super-resolution can introduce instability that affects quantitative interpretation. This indicates that the reliability of SR outputs for agronomic analysis strongly depends on the degree of downsampling and the preservation of inter-band relationships.
\end{enumerate}

\section{Conclusions}

We evaluated the applicability of eight super-resolution (SR) methods originally developed for RGB imagery to RGN multispectral crop images using both whole-image and leaf-level reconstruction analysis. Across both evaluation settings and scaling factors ($4\times$ and $16\times$), \modification{non-parametric Kruskal--Wallis tests followed by Dunn's post-hoc tests with Bonferroni correction confirmed that} transformer-based approaches significantly achieved the strongest overall performance  ($p < 0.001$), with \modification{HAT, HiT-SR, and SRFormer belonging to the same top-performing statistical tier with no significant pairwise differences}. Bicubic interpolation remained a competitive baseline under moderate degradation, although its ability to recover fine structural information remained limited \modification{and its performance became statistically comparable to Transformers at extreme $16\times$ scaling in several metrics}. The findings show that the reliability of vegetation index reconstruction depends on the specific index: soil-adjusted indices (SAVI and OSAVI) were reconstructed more consistently, whereas ratio-based indices (RVI and MSR) exhibited substantially greater sensitivity to spectral distortions, especially for $16\times$ scaling. Our results demonstrate that visual reconstruction quality is not a good proxy to assess SR methods for multispectral agricultural applications. Rather, preserving spectral relationships should be the determining factor for reliable quantitative analysis. The results indicate that transformer-based SR methods currently provide the most suitable balance between spatial enhancement and spectral accuracy for crop phenotyping tasks.

\medskip
\textbf{Supporting Information} \par 
Supporting Information is available from the corresponding author if required.

\medskip
\textbf{Acknowledgements} \par 
This work received support from the Agencia Nacional de Investigación y Desarrollo (ANID) under grant FONDECYT Iniciación, 11250090, and the Advanced Center of Electrical and Electronic Engineering (AC3E) grant Basal Project CIA250006 AC3E CIA250006.

\medskip
\textbf{Declaration of Generative AI and AI-assisted technologies in the
writing process}\par

The authors used generative AI tools, such as Google Gemini and ChatGPT, to refine the grammar and improve readability.

\medskip

%


\bibliographystyle{IEEEtran}
\bibliography{ref.bib}

\end{document}